%% file: main.tex
\documentclass[conference]{IEEEtran}
\IEEEoverridecommandlockouts

\usepackage{dsbda-style}
\usepackage{dsfont}
\usepackage{bbm}
\usepackage{array}
\newcolumntype{L}[1]{>{\raggedright\let\newline\\\arraybackslash\hspace{0pt}}m{#1}}
\newcolumntype{C}[1]{>{\centering\let\newline\\\arraybackslash\hspace{0pt}}m{#1}}
\newcolumntype{R}[1]{>{\raggedleft\let\newline\\\arraybackslash\hspace{0pt}}m{#1}}
\usepackage{paralist}
\usepackage{multirow} 
\usepackage{enumitem}
\usepackage{graphicx}
\usepackage{lipsum}

\usepackage{tabularx}
\usepackage{subcaption}

\usepackage{pifont}
\newcommand{\cmark}{\ding{51}}  
\newcommand{\xmark}{\ding{55}}  
\usepackage{bbding}

\usepackage[utf8]{inputenc} 
\usepackage[T1]{fontenc}    
\usepackage{hyperref}       
\usepackage{url}            
\usepackage{booktabs}       
\usepackage{amsfonts}       
\usepackage{nicefrac}       
\usepackage{microtype}      
\usepackage{xcolor}         

\author{
\IEEEauthorblockN{Jonatan Frank}
\IEEEauthorblockA{
\textit{University of Ulm}\\
Ulm, Germany \\
jonatan.frank@uni-ulm.de}
\and
\IEEEauthorblockN{David Richerby}
\IEEEauthorblockA{
\textit{University of Essex}\\
Colchester, UK\\
david.richerby@essex.ac.uk}
\and
\IEEEauthorblockN{Ansgar Scherp}
\IEEEauthorblockA{
\textit{University of Ulm}\\
Ulm, Germany \\
ansgar.scherp@uni-ulm.de}
}

\begin{document}

\title{Chimaera: A Mixture-of-Graph-Experts Architecture for Cross-Task and Cross-Dataset Graph Learning}

\maketitle

\begin{abstract}
Designing foundation models for graphs is challenging due to the irregular structure of graphs and the different sizes and characteristics of embeddings.
Chimaera integrates mixture-of-experts with graph foundation models (GFM). 
It integrates different GFM architectures, such as graph prompts and linear  GNN models.  
Large language models are used to generate embeddings, and experts can be trained and combined following different strategies, GFMs, embeddings, etc.
Furthermore, Chimaera extends existing linear GNNs to support link-level and graph-level tasks in addition to node-level tasks.
Empirical analyses are performed on same-task and cross-task experiments with node, link, and graph classification tasks using six benchmark text-attributed graph datasets.
The experiments demonstrate the effectiveness of Chimaera and its capabilities for transfer across tasks and datasets.
Further insights include the need to use both large and small language models to generate embeddings for the experts, a strong cross-task transferability of simple but effective linear GNNs, and using few samples only to provide strong results.
The source code: \url{https://github.com/ascherp/ChIMERA-dev}
\end{abstract}


\section{Introduction}
\label{sec:Introduction}

The success of foundation models~\cite{FM:Risk}\arxivonly{~in computer vision~\cite{FM:CV1,FM:CV2} and natural language processing~\cite{FM:GPT4, LLM:Gemini, LLM:Llama} inspired the development of foundation models specifically for graphs~\cite{GNN:GFM_Survey}. 
However, the significant successes in natural language processing} have not yet been replicated in graphs, because of their non-i.i.d. nature.
Learning from graphs is inherently more complex due to their irregular structure and the lack of a standardized vocabulary~\cite{GNN:Position_GFM}. 
Challenges faced by graph foundation models (GFMs)~\cite{GNN:GFM_Survey} include handling graph features of different kinds and sizes.
For example, a text-attributed graph may be encoded with sparse embeddings like TF-IDF or dense neural embeddings such as a BERT model~\cite{LLM:BERT}.
Additionally, there are two key characteristics one expects from a foundation model: homogenization~\cite{ML:Homogenization} and generalizability (also known as emergence)~\cite{ML:Emergent}.
\arxivonly{Homogenization refers to a model's applicability to different tasks.
Graph learning tasks are node classification, link prediction, and graph classification.
Generalizability is the model's ability to perform well on unseen datasets and tasks with an increasing amount of training data or parameters.} 

\begin{figure}
       \centering
  \includegraphics[width=0.4\textwidth]{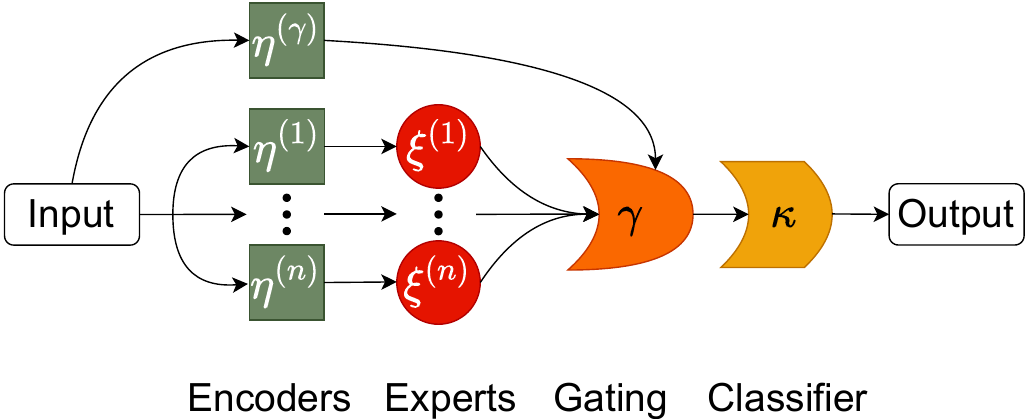}
  \caption{
  \arxivonly{Mixture-of-Graph-Experts configuration with Chimaera.} 
  We use $n+1$ input encoders $\eta$ (one encoder is from the original dataset).
  The input is combined via $n$ GNN experts $\xi$ in a gating function $\gamma$, followed by a classifier $\kappa$.}
    \label{fig:Chimaera}
   \end{figure} 
Even though graph neural networks (GNNs)~\cite{GNN:GCN, GNN:GIN, GNN:GAT, GNN:GraphSAGE} emerge as a specific family of graph learning methods that exhibit impressive performance in tasks such as node classification~\cite{GNN:GIN,GNN:Node1, GNN:CoGNN, GNN:Sylvie} and link prediction~\cite{GNN:GAT}, they are limited to the same feature and label spaces as they are trained on.
In other words, a GNN trained on a specific dataset for a certain task does not generalize to other datasets and tasks.
Hence, one has to try different models and configurations for each new dataset and task.
\arxivonly{A mixture-of-experts architecture can learn to combine different GNNs depending on their strengths and, therefore, can mitigate this problem. However, the dependency on the same feature and label spaces remains. A solution for the feature space dependency in text-attributed graphs is provided by large language models (LLMs), which transfer the text features of the node into the same embedding space. For the label space dependency, the solution can be found in prompting~\cite{GNN:PRODIGY} and linear GNNs~\cite{GNN:Trainless, GNN:GraphAny}.}

To solve these problems and create a GFM based on GNNs, we introduce Chimaera,\arxivonly{\footnote{``Chimaera'' (Greek mythology) is a creature part lion, goat, and serpent. In a similar way, we combine familiar elements to create something new.}} the first model that combines existing GFMs, such as prompting and linear GNNs,  with a mixture-of-experts (MoE) architecture.
In this context, we extend linear GNNs from node classification to include link-level and graph-level tasks.
Our model is illustrated in Figure~\ref{fig:Chimaera}, where the experts~$\xi^{(i)}$ represent a sequence of different GNNs operating on potentially different embeddings. 
We use different embeddings generated by neural encoders $\eta^{(i)}$, e.\,g., BERT~\cite{LLM:BERT}.
Consequently, Chimaera is fundamentally a GNN-based model but also functions as a GNN+LLM-based model for text-attributed graphs.
The gating function~$\gamma$ combines the output embedding of the experts and resembles a MoE. The classifier~$\kappa$ is a GFM that uses the combined embeddings for classification.
Thus, our model aims to generalize by integrating various experts and embeddings while ensuring homogenization through existing GFM methods. 
Chimaera supports node-, link-, and graph-level classification tasks and can transfer knowledge across tasks and datasets.
Thus, it is a GFM itself.

We experiment with various LLMs as encoders, GNNs as experts, gating functions, and GFMs as classifiers. 
We compare co-training with pre-training and evaluate our method in both same-task and cross-task test settings, following~\cite{data:GFMBenchmarks}.  
Co-training applies a trained model to similar datasets, while pre-training involves using trained models on datasets not included in the training phase.
Same-task models transfer knowledge within a specific task type, such as node classification; cross-task models focus on transferring knowledge between different tasks, \eg from node classification to graph classification.
Additionally, we use varying numbers of samples for the classification process using linear GNNs and prompting.

Using six benchmark datasets on four tasks (node classification, link prediction, reasoning, and graph classification), we observe that Chimaera performs well when trained on the same datasets and tasks, achieving scores similar to the baseline GNNs.
In cross-task or pre-training scenarios, Chimaera demonstrates strong generalization capabilities, achieving performance close to that of models directly trained on the target dataset in more than half of the cases. 
This highlights the robustness of Chimaera, especially given that, unlike the baselines, our model was not explicitly trained for those specific tasks or datasets. 
\arxivonly{In summary, our contributions are:
\begin{itemize}
\item We introduce Chimaera, a model that combines GFMs with MoE.
 As a feature by design, our model itself is a GFM that shows generalizability and homogenization.

 \item An analysis of complexity and an examination of a gating function across different contexts.

\item Experiments in $64$ settings on $6$ benchmarks with $4$ tasks demonstrate Chimaera's effectiveness in cross-task settings and transferring to new datasets with few training samples.

\end{itemize}
}

\arxivonly{
We summarize related work below.
Section~\ref{sec:Chimaera} introduces Chimaera.
The experimental apparatus is described in Section~\ref{sec:experimentalapparatus}.
An overview of the achieved results is reported in Section~\ref{sec:results} and discussed in Section~\ref{sec:discussion}, before we conclude.
}

\section{Related Work}
\label{sec:relatedwork}

There are recent comprehensive surveys of Graph Foundation Models (GFMs)~\cite{GNN:GFM_Survey,DBLP:journals/corr/abs-2505-15116}.
Below, we describe selected models relevant to this work.
Generally, GFM architectures can be divided into those based on GNNs, LLMs, or both~\cite{GNN:GFM_Survey,DBLP:journals/corr/abs-2505-15116}. 
PRODIGY~\cite{GNN:PRODIGY} is a pre-trained GNN-based GFM that can directly perform downstream classification tasks on unseen graphs using in-context learning and graph prompts.
GraphAny~\cite{GNN:GraphAny} is a GNN-based model that combines different linear GNNs~\cite{GNN:SGC} with an attention module. 
The linear GNNs need no training and are adapted for each graph using an analytical solution. 
Only the attention module in GraphAny is trained. 
TrainlessGNN~\cite{GNN:Trainless} is a another linear GNN adapted for graph learning tasks.
The LLM-based GLM~\cite{LLM:GraphLM} uses the encoder-decoder language model T5~\cite{LLM:T5}, and modifies the adjacency matrix to support graphs, and converts text-attributed graphs into Levi graphs. 
GPT4Graph~\cite{LLM:GPT4Graph} uses a graph description language and the language capabilities of GPT4~\cite{FM:GPT4} to process graph data.
GNN+LLM-based GFMs combine features of GNNs and LLMs to create synergies. 
For example, OFA~\cite{GNN:OFA} uses LLMs to improve embeddings for GNNs by incorporating graph features and text prompts.
Another model, GOFA~\cite{GNN:GOFA},  interleaves randomly initialized GNN layers into a frozen pre-trained LLM, allowing for a natural integration of semantic and structural modeling capabilities.
These methods are limited to textual features within the graph or require textual prompts.

A Mixture-of-Experts (MoE)~\cite{ML:MoEOrigin} combines a set of experts, \eg neural networks, using a mixture function and returns a distribution over the experts.
Fedus \etal~\cite{LLM:SwitchT} suggest using only one expert at a time and encourage a balanced workload among experts by incorporating an auxiliary loss~\cite{LLM:MoE, LLM:Mesh, LLM:Giant}.
To enhance the scaling power of MoE, GRIN incorporates sparse gradient estimation for expert routing and configures model parallelism~\cite{LLM:GRIN}.
GMoE~\cite{GNN:GMoE} is based on GNNs, and enhances the network's adaptability to diverse training graphs with minimal computational overhead. 
DA-MoE~\cite{GNN:DA-MoE} is a graph MoE that uses a GNN, in contrast to the linear projections used by GMoE, to better capture complex patterns and dependencies within the data.
Previous work, e.\,g., \cite{GNN:MoLP}, has primarily focused on using MoE to enhance performance on specific graph-related tasks, such as link prediction. In contrast, our research aims to use MoE to transfer knowledge across multiple tasks and datasets.

\section{Chimaera MoE-Architecture for GFMs}
\label{sec:Chimaera}

Given a graph $G = (V, E)$ with nodes $V$ and edges $E$, we represent its features as
$(\mX,\mA)$, where $\mX \in \mathbb{R}^{|V| \times c}$ is the feature matrix of the nodes and $\mA \in \mathbb{R}^{|V|\times |V|}$ is the adjacency matrix.
The set of classes is denoted by $Y$, which specifies a task-dependent aspect of the graph.
We consider four tasks.
In node classification, each node gets a label from~$Y$.
For link prediction, a pair of nodes is predicted to have a $1$ if the edge between these two nodes exists or $0$ otherwise, \ie $Y = \{0,1\}$.
In reasoning, the edges get a label from~$Y$.
In graph classification, we assign a label to each graph~$G$.

As shown in Figure~\ref{fig:Chimaera}, the input is first encoded by the encoders $\eta^{(i)}$ to create the embeddings $\mX^{(i)}\in \mathbb{R}^{|V|\times c}$ before passing $\mX^{(i)}$ to the expert $\xi^{(i)}$, \eg a GNN.
The experts themselves provide their embeddings $\mH^{(i)}\in \mathbb{R}^{|V|\times d}$  to the gating function $\gamma$.
All experts have the same output dimension, but their input dimension my differ depending on the encoder.
A skip connection is provided from the input to the gating function $\gamma$ by the encoder $\eta^{(\gamma)}$. 

The gating function $\gamma$ combines the embeddings $\mH^{(1)},\dots,\mH^{(n)}$ from the experts and the embedding $\mX^{(\gamma)}$ created by the encoder $\eta^{(\gamma)}$ to create an embedding $\mF\in \mathbb{R}^{|V|\times d}$. 
Then a classifier $\kappa$, a GFM, returns a prediction $\mZ\in \mathbb{R}^{|V|\times |Y|}$ for $\mF$.
Given the total of $n+1$ embeddings $\mX$ from the $n+1$ encoders $\eta$, a gating function $\gamma$, and a classifier $\kappa$, we define Chimaera as 
  $\chi(\mX^{(\gamma)},\mX^{(1)},\dots,\mX^{(n)}) = \kappa\big(\gamma\big(\mX^{(\gamma)},\xi^{1}(\mX^{(1)}),\dots,\xi^{n}(\mX^{(n)})\big) \big)$.  

The choice of encoders, experts, gating functions, and classifiers is an optimization aspect of Chimaera that is not typically found in other GNNs or GFM architectures.
We introduce the implementations of encoders, experts, gating functions, and GFMs.
The GFMs we consider include linear GNNs~\cite{GNN:GraphAny, GNN:Trainless} and the prompt-based PRODIGY~\cite{GNN:PRODIGY}.
Subsequently, we explain how to train Chimaera effectively.

\paragraph{Encoders}
We use language models as encoders to create node embeddings for our text-attributed graphs.
Our encoders are BERT~\cite{LLM:BERT}, Sentence-BERT~\cite{LLM:Sentence} (ST), E5~\cite{LLM:E5}, Llama2 7B, and Llama2 13B~\cite{LLM:Llama}. 
\arxivonly{We use LLMs as encoders, but any model capable of producing node embeddings could also be integrated into Chimaera.}

\paragraph{Experts}
Experts in Chimaera can differ in architecture, size, and parameters.
Their objective is to create embeddings $\mH^{(i)}$ by enhancing the input embeddings $\mX^{(i)}$ with additional information or refining them. 
We use the following GNNs as our expert models: GCN~\cite{GNN:GCN}, GIN~\cite{GNN:GIN}, and GAT~\cite{GNN:GAT}. 
\arxivonly{We use these GNNs as they show state-of-the-art performance when optimized properly~\cite{GNN:Strong1,data:pitfalls}.}

\paragraph{Gating Functions}
For simplicity, we refer here to the embedding $\mX^{(\gamma)}$ created by the encoder $\eta^{(\gamma)}$ for the gating function $\gamma$ as input $\mX$.
The gating function combines the experts' output embeddings.
One straightforward approach for a gating function to calculate an embedding matrix $\mF$ of the nodes is to calculate the average of the embeddings $\mH^{(1)},\dots,\mH^{(n)}$ (recall that these embeddings have the same dimension). Alternatively, we can concatenate the embeddings and feed them through an MLP to generate the embedding~$\mF$. We refer to these methods as \textit{mean} and \textit{MLP}, respectively.
The mixture functions used in MoE approaches are particularly well suited to serve as gating functions.
The original MoE formulation~\cite{ML:MoEOrigin} combines a set of experts $\xi^{(1)},\dots, \xi^{(n)}$ using a mixture function~$Q$ (\eg an MLP) that outputs a distribution, represented as a vector, over the experts given the input~$\mX$.
Chimaera uses a mixture function as a gating function with 
$\gamma = \sum^n_{i=1}Q(\mX)_i\cdot \xi^{(i)}(\mX^{(i)})$,
where $Q(\mX)_i$ denotes the $i$th item in the output vector of $Q(\mX)$.
Thus, $Q$ determines the weighting of the experts appropriate to a given input.
\textit{GMoE}~\cite{GNN:GMoE} employs a noisy top-$k$ gating design~\cite{LLM:MoE} and is defined as
 $Q(\mX) = \text{Softmax}(\text{TopK}(\mX\mW_g +
  \epsilon\, \text{Softplus}(\mX\mW_n)),k))$.
Here, $k$ is the number of selected experts, 
$\epsilon \sim \mathcal{N}(0,1)$ is standard Gaussian noise, and  
$\mW_g\in \mathbb{R}^{c \times n}$ and $\mW_n \in \mathbb{R}^{c \times n}$ are learnable weights.
\textit{DA-MoE}~\cite{GNN:DA-MoE} uses a structure-based gating network instead of linear projection to obtain the scores for each expert: 
    $Q(\mX) = \mathrm{Softmax}(\mathrm{TopK}(T(\mX) + \epsilon \, \text{Softplus}(T(\mX)),k))$ 
    with
    $T(\mX) = \sigma ((1+\alpha) \mX + \mA \mX )$, where 
$\alpha$ is a learnable parameter adjusting the contribution of the
node’s own feature and $\sigma$ is a two-layer fully connected
neural network with a nonlinear activation function applied
between the layers.

\paragraph{Classifiers}
As classifiers, we employ the linear GNN models LinearGNN from 
GraphAny~\cite{GNN:GraphAny} and TrainlessGNNs~\cite{GNN:Trainless}, and use the prompt-based PRODIGY~\cite{GNN:PRODIGY}.
To extend Linear GNNs to all task types, we convert the original graph $G$ into a task graph $G_T = (V_T, E_T)$. 
The nodes $V_T$ represent entities in the original graph $G$.
Depending on the task, $V_T$ represents the original nodes for the node-level tasks, edges for link-level tasks, and graphs for the graph-level tasks.
We use the embeddings in $\mF$ of the gating function to create the feature matrix $\mP$ of the nodes $V_T$  depending on the task. 
For node-level tasks, $\mP = \mF\in \mathbb{R}^{|V|\times d}$, and for a link-level task with $m$ edges, we concatenate the respective node embeddings in $\mF$ to create $\mP \in \mathbb{R}^{m\times 2d}$. 
In graph-level tasks with $m$~graphs, we use mean as a pooling function to create $\mP\in \mathbb{R}^{m\times d}$ from $\mF$.
We use only $\mP$ for classification and do not update the features of the task graph $G_T$, so we set $E_T = \emptyset$.
The labeled nodes in the task graph $G_T$ are represented by $V_L$, along with their corresponding embeddings, denoted as $\mP_L$. These labeled nodes have the same label as the entities in the original graph they represent.
The prediction of a linear GNN is $\mZ=\mP\mW$.
The learnable weight matrix $\mW \in \mathbb{R}^{e\times |Y| }$ has the dimensionality $e=2d$ for link-level tasks and $e=d$ for the other tasks.

\arxivonly{
We distinguish between two types of linear GNNs to establish the weight matrix $\mW\!$.
The weight matrix of a LinearGNN from GraphAny~\cite{GNN:GraphAny} is determined by a closed-form solution $\mW=\mP_L^+\mY_L \in~\mathbb{R}^{e\times |Y|}$, 
where $\mP_L^+$ is the pseudo-inverse of the embeddings of the labeled nodes in $V_L$, and $\mY_L\in \mathbb{R}^{|V_L|\times |Y|}$ are the label vectors of the nodes in $V_L$.
In contrast, TrainlessGNNs~\cite{GNN:Trainless} use virtual class nodes for creating the weight matrix. The virtual class nodes are connected to the nodes in $V_L$ that belong to the same class via edges with weight~$1$. 
The virtual nodes are connected to nodes belonging to different classes by edges with a weight defined by the hyperparameter $\omega$.  
This results in $\mB_L \in \{0,1\}^{|V_L|\times |Y|}$.
The matrix $\mB_L$ acts as the incidence matrix between virtual class nodes and $V_L$.
One round of message-passing is applied~\cite{GNN:Trainless} to obtain the weight matrix $\mW \in \mathbb{R}^{e \times |Y|}$: 
$\mW^T = ( \mB-\frac{\omega}{|Y|} \mathbbm{1} )^T\mP_L,$
where $\mathbbm{1}$ is the all-ones matrix of size $|V_L|\times |Y|$.

In the context of graph prompting, a prompt set $S$ consists of $k$ example graphs per class.
The objective is to use the prompt set $S$ to classify the entities in a query set $Q$. 
PRODIGY applies this approach for classification in the graph domain~\cite{GNN:PRODIGY}.
It uses a prompt graph as a unified representation for $k$-shot prompts across $|Y|$ labels with $o$ queries.
A prompt graph is composed of data graphs and a task graph.
Each entity from the prompt set $S$ and the query set $Q$ is represented by a data node in the task graph.
The data graphs are used to compute the embeddings of these data nodes. 
In Chimaera, the embeddings $\mP$ of the data nodes are created using the embeddings in $\mF$ from the gating function.
For graph tasks, the data node embeddings are obtained through mean pooling, while for node tasks, the respective node embeddings are used.
In the case of link tasks, the embeddings of the nodes forming the edge and the max pooling over all node embeddings~\cite{GNN:MaxP} are concatenated with each other.
With an additional linear projection
layer, the embedding sizes are converted back to $d$.
Each possible class $y$ is represented by a class node $v_y$ in the task graph.  
The embeddings $\mP_y$ of the class nodes can be initialized with a random Gaussian or with additional information about the class. However, we initialize all entries with zero to ensure consistent conditions for our experiments. 
In total, a task graph contains $|Y|\cdot k+o$ data nodes and $|Y|$ class nodes. 
For the prompts, one connects each data node to all the class nodes and vice versa.
The edges are labeled with ``True'' if the prompt example belongs to the class of the label, and otherwise ``False''.
 For the query set, one adds single-directional edges from all class nodes to each data point in the query set, which is labeled with ``?''.
An attention-based GNN~\cite{GNN:PRODIGY} is then applied to the task graph to produce an updated representation $\mP_Q$ of the query nodes and $\mP_C$ of the class nodes.
The classification is done by using the cosine similarity $\mathrm{sim}$ between $\mP_Q$ and $\mP_C$, \ie 
$\mZ = \mathrm{sim}(\mP_Q,\mP_C)\,.$
}

\paragraph{Training Chimaera}
\label{sec:CreateCh}
Chimaera is trained stepwise along the components shown in Figure~\ref{fig:Chimaera}. 
\arxivonly{For a task on a dataset divided into training, validation, and test sets, we construct and train Chimaera as follows.}
We first train the experts using a simple classifier on the dataset with the embeddings from the respective encoder and validate them on the validation set.
Subsequently, the classifiers are removed, and the experts' parameters are frozen. 
The validation results are used to select the experts and their encoders.
We propose three strategies for using GCN, GAT, and GIN as experts.
The first strategy, the \textit{multi-GNN strategy}, selects the best GCN, GAT, and GIN with their encoders based on their validation performance.
Second, the \textit{multi-encoder strategy} chooses the best GNN with its encoder according to its validation results and combines it with two GNNs that are the same GNN but use different encoders.
Third, the \textit{multi-layer strategy} again selects the best GNN and its encoder and combines it with two GNNs that differ from it solely in the number of layers and have the same encoder.

The chosen expert and encoder pairs are then combined using a gating function.
The gating function has a simple classifier on top and is trained. Afterward, the classifier is discarded, and the parameters are frozen. 
Then, the gating function is used to generate input for a chosen GFM as a classifier.
The GFM is trained, and then its parameters are frozen. 

The training of Chimaera reduces both the overall training time and complexity when compared to training all components simultaneously. This approach reduces the hyperparameter space required for grid search and enables the parallel training of the experts.
Additionally, the training time for each individual component remains unaffected since the hyperparameters of the underlying components are kept frozen. 

\arxivonly{
\paragraph{Complexity Analysis}
For the complexity analysis, we look into the components of Chimaera.
The inference complexity of an $\ell$-layer GNN expert is $\mathcal{O}(\ell\, (|E|d+nd^2))$, where $|E|$, $n$, and $d$ are the number of edges, the number of nodes, and the hidden dimension, respectively.

Chimaera can use various gating functions with differing complexities.
For the gating using mean, no additional overhead is added because it has a linear complexity of $\mathcal{O}(d)$. The MLP gating, GMoE, and MoE use an MLP for gating with complexity $\mathcal{O}(d^2)$.
It was also shown that GMoE~\cite{GNN:GMoE} brings negligible overhead on the inference cost compared with its GNN counterpart.
For DA-MoE, an $\ell$-layered GNN is used for gating, which has a complexity of $\mathcal{O}(\ell\, (|E|d+nd^2))$.

Given $m$ experts with $\ell_1, \dots, \ell_m$ GNN layers, respectively, the inference complexity of our gating is $\mathcal{O}(\sum_{i=1}^m \ell_i\,(|E|d+nd^2))$.
Overall, the cost of gating on the inference cost is negligible when the gates and experts can be run simultaneously.

A linear GNN has pre-processing complexity of $\mathcal{O}(|E|)$, optimization complexity of  $\mathcal{O}(|V_L|)$, and inference of $\mathcal{O}(|V|)$. Therefore, with a linear GNN, the complexity of Chimaera is $\mathcal{O}(\sum_{i=1}^m \ell_i\, (|E|d+nd^2))$.
PRODIGY uses a GNN on the task graph with $|Y|\, k +o$ data nodes and $|Y|$ class nodes. We approximate the complexity of creating the task graph and the inference on it with an $\ell_0$-layer GNN as $\mathcal{O}(\ell_0\, (|E|d+nd^2))$ because, in general, the task graph is much smaller than the original graph. 
The resulting Chimaera inference complexity is  $\mathcal{O}(\sum_{i=0}^m \ell_i\, (|E|d+nd^2))$.}

\section{Experiments}
\label{sec:experimentalapparatus}

\arxivonly{
\begin{table*}[h]
    \centering
        \caption{Characteristic of TAGLAS datasets~\cite{data:TAGLAS}.}
    \label{tab:data_TAGLAS}
\begin{tabular}{llrrrlcc}
\textbf{Task type}&\textbf{Dataset} &\textbf{Avg. $|V|$} &\textbf{Avg. $|E|$ }& \textbf{\# graphs} &\textbf{Domain} &\textbf{Split (train/val/test)} & \textbf{\# classes} \\[1ex]\hline\\
        Node-level &Cora& 2\,708 & 21\,112 & 1 & Co-citation &$140/500/2\,068$&7 \\
        &PubMed & 19\,717 & 44\,338  & 1 & Co-citation &$60/500/19\,157$&3 \\
        &WikiCS& 11\,701 & 216\,123  & 1 & Wikipedia page &$580/1\,769/5\,847$& 10 
        \\[2ex]
        Link-level & Cora & 2\,708 & 21\,112 &  1 & Co-citation &$17\,944/1\,056/2\,112$&2 \\
        &PubMed& 19\,717 & 44\,338 &  1 & Co-citation &$150\,700/8\,866/17\,730$&2 
        \\
        &WN18RR & 40\,943 & 93\,003 &  1 & Knowledge graph &$86\,835/3\,034/3\,134$&11 \\[2ex]
        Graph-level & BBBP & 24.06 & 51.91 & 2\,039 & Molecular &$1\,631/204/204$&2  \\
        &BACE & 34.09 & 73.72 &  1\,513 & Molecular &$1\,210/151/152$&2
    \end{tabular}
\end{table*}
}

\paragraph{Datasets}
We use six datasets from TAGLAS~\cite{data:TAGLAS}, an atlas of text-attributed graph datasets and benchmarks.
\arxivonly{An overview of the dataset characteristics is provided in Table~\ref{tab:data_TAGLAS}.}
For node classification, we use the datasets Cora, PubMed~\cite{GNN:OFA, data:Graphllm}, and WikiCS~\cite{data:mernyei2020wiki,GNN:OFA}.
In link-level tasks, we distinguish between link prediction, which classifies whether an edge exists, and reasoning, which aims to label existing edges.
We use the Cora and PubMed datasets~\cite{GNN:OFA, data:Graphllm} for link prediction. For reasoning, we use the WN18RR dataset~\cite{GNN:OFA}. 
Finally, for graph-level tasks, we include the BBBP and BACE datasets~\cite{data:gimlet}.

\paragraph{Procedure}
We use several pre-trained models to obtain embeddings from text-attributed graphs, including BERT~\cite{LLM:BERT}, Sentence-BERT~\cite{LLM:Sentence}, E5~\cite{LLM:E5}, Llama2 7B, and Llama2 13B~\cite{LLM:Llama}. 
\arxivonly{The embeddings generated by BERT and Sentence-BERT have dimension $c=768$, while those created by E5 are $c=1,024$. 
The largest embeddings are created by Llama2 7B with $c=4,096$ and Llama2 13B with $c=5,120$.}
For the experts, we employ GIN~\cite{GNN:GIN}, GAT~\cite{GNN:GAT}, and GCN~\cite{GNN:GCN}. 
All experiments are conducted on an NVIDIA H100 GPU with 80GB. 

\textit{Main Experiments:}
We train various combinations of embeddings and experts using classification heads on the training data, in which the encoders are frozen.
The combinations are aggregated using one of our gating functions.
To select the experts for the next steps, we apply one of our three strategies.
These strategies are the multi-GNN strategy, the multi-encoder strategy, and the multi-layer strategy (see Section \ref{sec:Chimaera}).
After determining the experts, we freeze their parameters and train the gating function using the training data.
Subsequently, we use one of our gating functions or the best GNN model as input for the GFM, \ie the classifier $\kappa$ (see Figure~\ref{fig:Chimaera}).
As classifiers, we use LinearGNN~\cite{GNN:GraphAny}, TrainlessGNN~\cite{GNN:Trainless}, and PRODIGY~\cite{GNN:PRODIGY}. 
Only the prompt-based PRODIGY is trainable. 
Thus, it is the only model trained on the training data.
See also the explanation of GFMs as classifiers in Section~\ref{sec:Chimaera}. 
We use trained experts and the gating function as our baselines.
GNNs serve as valid baselines for graph-related tasks~\cite{GNN:Strong1}.  A model that has been explicitly trained on a specific task and dataset naturally serves as a baseline for another model that has not been trained on that task or dataset, particularly when it needs to transfer knowledge between different tasks or datasets.

\textit{Experiments with Full Chimaera and Using PRODIGY as Classifier:}
First, we train Chimaera with a linear GNN and parameters unfrozen. 
The full Chimaera is trained after our regular training procedure. 
Second, we train Chimaera with PRODIGY, similar to our training procedure except that we train PRODIGY for node-, link-, and graph-level tasks by combining the loss of these tasks in each training step.

For the testing procedure, we follow~\cite{data:GFMBenchmarks}, and vary the number of shots per class. 
We use either $3$, $10$, or $20$ shots per class,  with a maximum of $10$ for WikiCS due to the limited number of examples in WikCS. 
We test five times with different shots and random seeds, averaging the results over these five runs. 
We explore the same-task and cross-task test settings
and differentiate between co-training and pre-training, as described in the introduction. 
For each combination of settings, we evaluate the models' performance on the specific test set, using accuracy for tasks with more than two classes and the ROC curve (AUC) for binary tasks.

\textit{Experiments on Gating Functions:}
To investigate how MoE behaves for tasks and data for which it has not been trained, we track the scores of the gating function of DA-MoE with $k=3$ for each expert for each tested sample. 
This analysis is exclusive to DA-MoE because it uses the graph's structure in its gating function. 
As a result, we obtain a distribution of scores for each expert, which we can further categorize by class and training data. We employ three different strategies and compare their outcomes. Our objective is to identify variations based on class, expert, and the strategy implemented.

\paragraph{Hyperparameter Optimization}
For the hyperparameter optimization, we use grid search.
We use GIN, GAT, and GCN as experts with input generated by either BERT, Sentence-BERT, E5, Llama2 7B, or Llama2 13B. The encoders are not trained and are only used for inference.
The experts are trained for $50$ epochs on the training set, employing early stopping with a patience of $5$ epochs. 
We conduct a grid search over all combinations of embedding size $\in \{64,128,256\}$, number of layers $\in \{1,2,3\}$, dropout $\in \{0,0.2\}$, and weight decay $\in \{0,0.01\}$. 
For TrainlessGNNs, we optimize $\omega \in  \{-1, 0, 0.001, 0.01, 0.1, 1\}$, which is an edge weight modeling the influence of classes of which the node under consideration is not a member.
For node and graph classification tasks, we use a $1$-layer MLP as the classification head, while for link tasks, we employ a $2$-layer MLP. 
All models are optimized using the Adam optimizer\arxivonly{~\cite{Ml:Adam}} with a learning rate of $0.01$.

MoE employs a $2$-layer MLP with an embedding size matching the experts. GMoE uses a $2$-layer MLP with the same embedding size and varies $k$ between $2$ and $3$. 
DA-MoE selects the best $k$ experts for its gating function, with $k\in\{2,3\}$. It uses the best GNN for its mixture function.
The MLP as a gating function takes three times the embedding size of the experts as input and consists of two layers with a hidden size equal to the embedding size.
We train the gating functions for $50$ epochs, again using early stopping with a patience of $5$ epochs.
These models are optimized with the Adam optimizer\arxivonly{~\cite{Ml:Adam}}, learning rate $0.01$, and weight decay~$0$. 

We train PRODIGY for Chimaera on one task of one dataset using the procedure from~\cite{GNN:PRODIGY}, using three shots per example. 
Training is conducted with $2,000$ batches. 
We use the standard values from the PRODIGY implementation: Adam optimizer\arxivonly{~\cite{Ml:Adam}}, learning rate $0.001$, and weight decay $0.001$.

For training the whole of Chimaera with a linear GNN, we train it similarly to PRODIGY, but for $500$ batches. 
For TrainlessGNNs, we set $\omega = 0$. For the multi-task PRODIGY, in addition to the training dataset, we incorporate the Cora dataset to introduce node-level and link-level tasks, as well as the BACE dataset to add a graph-level task to the training process. If a task is already included in the training dataset, we do not use the additional dataset for that specific task.  
Each batch consists of samples for each task. 
All tasks are executed separately during each training step, and their losses are combined. 
All models are optimized with the Adam optimizer\arxivonly{~\cite{Ml:Adam}}, learning rate $0.001$, and weight decay $0.001$.

\section{Results}
\label{sec:results}

\begin{table}
    \centering
        \caption{
Best test accuracy/AUC of the experts, gating functions, and Chimaera. In training, pre-/co- indicate pre-/co-training, S/X\,=\,same/cross-task.
\textbf{--} means no suitable dataset.}
\label{tab:CompChimaera}
    \small
    \setlength\tabcolsep{4pt}
\begin{tabular}{l|ccc|ccc}
\textbf{Train\!\!}&\multicolumn{3}{c|}{\textbf{Accuracy}}&\multicolumn{3}{c}{\textbf{AUC}}\\& Expert & Gating & Chimaera & 
Expert & Gating & Chimaera \\\hline
&\multicolumn{3}{c|}{Cora (node)}&\multicolumn{3}{c}{Cora (link)}\\
co-S  &0.8&0.82&0.81$\pm$0.00&0.99&0.99&0.60$\pm$0.01\\
co-X&n/a&n/a&0.76$\pm$0.00&n/a&n/a&0.68$\pm$0.00\\
pre-S &n/a&n/a&0.67$\pm$0.00&n/a&n/a&0.69$\pm$0.00\\
pre-X&n/a&n/a&0.67$\pm$0.00&n/a&n/a&0.71$\pm$0.00\\[0.75ex]
&\multicolumn{3}{c|}{PubMed (node)}&\multicolumn{3}{c}{PubMed (link)}\\
co-S  &0.79&0.80&0.80$\pm$0.00&0.98&0.99&0.90$\pm$0.00\\
co-X&n/a&n/a&0.74$\pm$0.00&n/a&n/a&0.82$\pm$0.04\\
pre-S &n/a&n/a&0.61$\pm$0.00&n/a&n/a&0.75$\pm$0.01\\
pre-X&n/a&n/a&0.70$\pm$0.00&n/a&n/a&0.89$\pm$0.00\\[0.75ex]
&\multicolumn{3}{c|}{WikiCS (node)}&\multicolumn{3}{c}{BBBP (graph)} \\
co-S &0.82&0.82&0.81$\pm$0.00&0.65&0.64&0.65$\pm$0.00\\
co-X &n/a&n/a&--&n/a&n/a&--\\
pre-S &n/a&n/a&0.48$\pm$0.02&n/a&n/a&0.64$\pm$0.00\\
pre-X&n/a&n/a&0.53$\pm$0.01&n/a&n/a&0.61$\pm$0.00\\[0.75ex]
&\multicolumn{3}{c|}{WN18RR (link)}&\multicolumn{3}{c}{BACE (graph)} \\
co-S &0.73&0.79&0.40$\pm$0.00&0.77&0.76&0.74$\pm$0.00\\
co-X &n/a&n/a&--&n/a&n/a&--\\
pre-S &n/a&n/a&--&n/a&n/a&0.63$\pm$0.04\\
pre-X &n/a&n/a&0.40$\pm$0.00&n/a&n/a&0.71$\pm$0.01
\end{tabular}
\end{table}

\begin{figure}
     \centering
 \begin{subfigure}[b]{0.49\textwidth}
       \centering
  \includegraphics[width=1\textwidth]{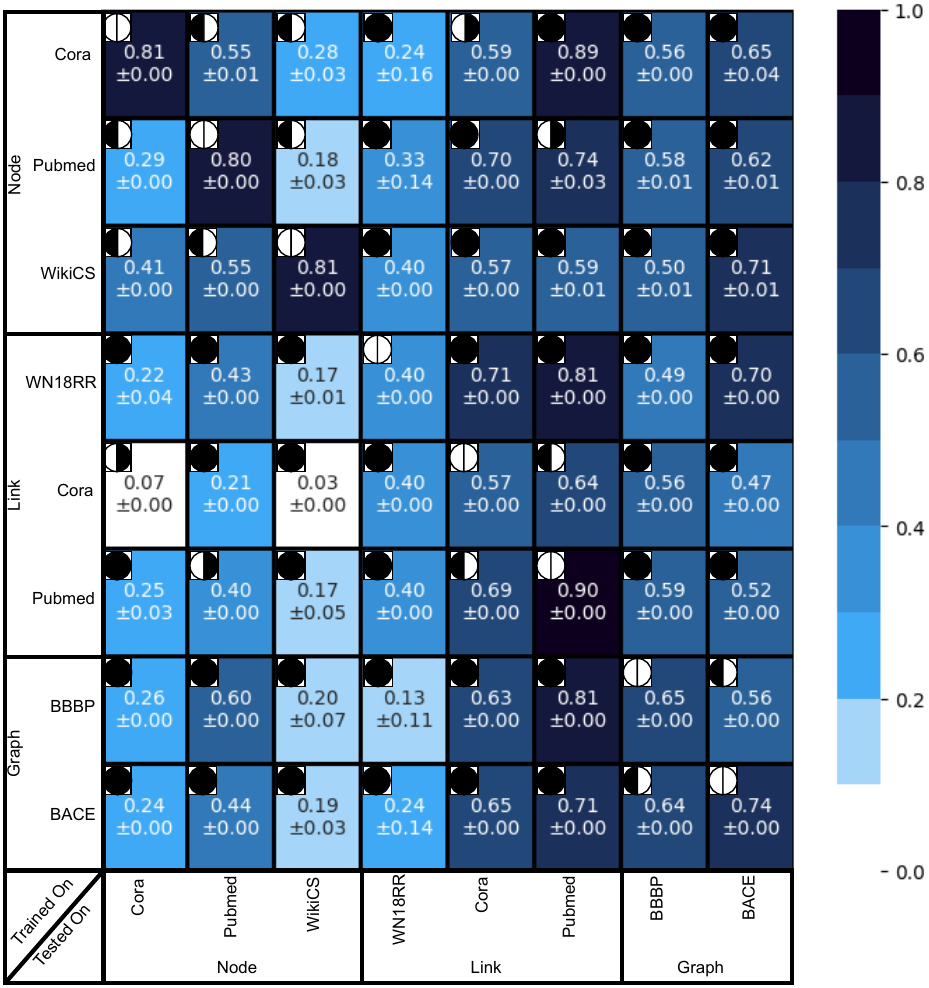}
  \caption{Chimaera with PRODIGY}
    \label{fig:ProdigyHeat}
   \end{subfigure} 
\begin{subfigure}[b]{0.49\textwidth}
       \centering
  \includegraphics[width=1\textwidth]{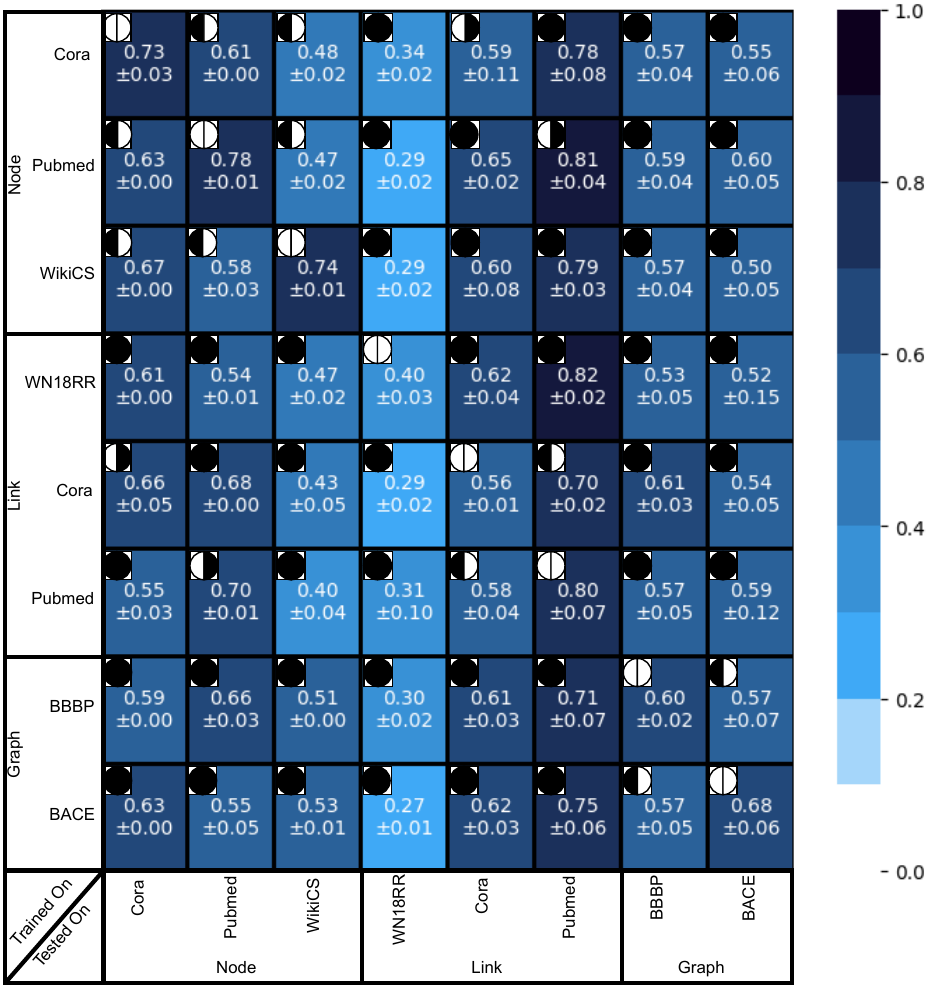}
  \caption{Chimaera with LinearGNN}
    \label{fig:linearHeat}
   \end{subfigure}  
    \caption{Chimaera's performance averaged over five runs using accuracy for tasks with more than two classes and AUC for binary tasks.  
    The circles in the tiles indicate the experimental setting. If a \textit{left semi-circle is white, it denotes co-training}; if it is \textit{black, it indicates pre-training}. 
    The \textit{right semi-circle is white for the same-task setting}, and \textit{black for cross-task}.}
        \label{fig:heatmaps}
\end{figure}

Our experiments show that the effect of the number of shots on PRODIGY in Chimaera is minimal.
Additionally, the performance of linear GNNs tends to improve with a higher number of shots. 
Unfreezing the parameters and learning a multi-task PRODIGY does not enhance Chimaera's overall performance.
The fully trained Chimaera with linear GNNs performs less well than training step-by-step and freezing. 
The results for Chimaera with PRODIGY and multi-training are mostly better for unseen datasets than for its variant with single-task training. Nevertheless, Chimaera with linear GNNs still outperforms it for unseen datasets.

Table~\ref{tab:CompChimaera} shows how the best Chimaera (in the co/pre-training and same/cross-task settings) performs compared to the best experts and gating functions, with the latter always being trained on the dataset and task it is tested on (always co-training and same-task).
Chimaera achieves performance comparable to the trained expert and gating functions for node classification in Cora and PubMed, link prediction in PubMed, and graph classification in BBBP and BACE independent of the test settings. 

The heatmaps in Figure~\ref{fig:heatmaps} show the best performance for Chimaera with PRODIGY and with LinearGNN (from GraphAny).
The heatmap for Chimaera with TrainlessGNN is similar to that of LinearGNN and is therefore omitted.
For cross-task and co-training, the linear GNNs effectively transfer knowledge between node classification and link prediction tasks.
When we apply pre-training on the same task, the linear GNNs perform well. 
In pre-training and cross-task scenarios, the linear GNNs excel in transferring knowledge from all tasks to link prediction and graph classification.

\begin{table}
\small
    \centering
        \caption{Test accuracy and AUC of all MoE combinations. \textit{Underlined}: MoE models better or equal to the best GNN. 
        \textit{Bold}: best MoE. 
        DA-MoE and GMoE with $k = 2$ omitted as $k = 3$ is either better or similar. N\,=\,Node, L\,=\,Link, G\,=\,Graph.}
    \label{tab:GateTest}
    \setlength\tabcolsep{.5pt}
\begin{tabular}{lcccccccc}
\multirow{3}{*}{\textbf{\shortstack[l]{Gating\\Function}}}&\multicolumn{4}{c}{\textbf{Accuracy}}& \multicolumn{4}{c}{\textbf{AUC}}\\
& Cora & PubMed &  WikiCS & WN18RR &  Cora &  PubMed & BBBP & BACE \\
&(N)&(N)&(N)&(L)&(L)&(L)&(G)&(G)\\ \hline\\
\multicolumn{9}{c}{\textbf{Multi-GNN strategy}} \\
mean&0.788&0.787&\underline{\textbf{0.816}}&\underline{0.751}&0.987&0.942&0.627&0.753\\
MLP&0.782&0.786&0.810&\underline{\textbf{0.791}}&0.980&0.908&0.634&0.436\\
MoE&0.773&0.768&0.762&\underline{0.750}&0.987&\underline{0.987}&0.617&0.755\\
GMoE&0.779&0.777&0.784&\underline{0.754}&\underline{\textbf{0.990}}&0.961&0.618&\textbf{0.758}\\
DA-MoE&0.787&0.793&0.806&\underline{0.736}&0.986&0.939&\textbf{0.642}&0.732\\[0.75ex]  
\multicolumn{9}{c}{\textbf{Multi-encoder strategy}} \\
mean&0.773&0.793&0.811&0.720&0.988&\underline{0.983}&0.611&0.745\\
MLP&0.756&\underline{0.798}&0.803&\underline{0.762}&0.987&\underline{0.982}&0.626&0.730\\
MoE&0.681&0.791&0.788&\underline{0.729}&\underline{0.989}&\underline{0.984}&0.614&0.740\\
GMoE&0.756&0.785&0.798&\underline{0.728}&\underline{0.989}&\underline{0.984}&0.620&0.745\\
DA-MoE&0.778&\underline{\textbf{0.800}}&0.810&0.716&\underline{\textbf{0.990}}&\underline{0.982}&0.623&0.707\\[0.75ex]
\multicolumn{9}{c}{\textbf{Multi-layer strategy}} \\
mean&\underline{\textbf{0.818}}&0.774&\underline{\textbf{0.816}}&\underline{0.744}&0.987&\underline{0.987}&0.621&0.712\\
MLP&0.762&0.776&\underline{\textbf{0.816}}&\underline{0.770}&0.987&\underline{0.988}&0.630&0.379\\
MoE&0.754&0.774&0.780&\underline{0.745}&\underline{0.989}&\underline{0.988}&0.631&0.658\\
GMoE&\underline{0.810}&0.772&0.813&\underline{0.749}&0.987&\underline{0.988}&0.625&0.733\\
DA-MoE&\underline{0.809}&0.778&\underline{0.815}&\underline{0.744}&0.987&\underline{\textbf{0.989}}&0.632&0.709
\end{tabular}
\end{table}

No variant of Chimaera had one gating function and one strategy that were best in all tested scenarios. 
However, in more than $93\%$ of our test cases, a gating function is the best input for a GFM.
The number of shots can influence which gating function is the most suitable input for an expert. 
Table~\ref{tab:GateTest} displays how well each combination of gating function and strategy performs directly on the test sets compared to the best GNNs listed in Table~\ref{tab:ModelsAndEncodersOverviewAcc}. 
Overall, the gating function enhances test accuracy or AUC for all tasks except those at the graph-level.
 For both GMoE and DA-MoE, using $k=3$ yields better results than $k=2$.  

\begin{table}
\small
    \centering
        \caption{
    Test accuracy and AUC for all encoders and models with the best validation accuracy and AUC. \#L = \#Layers}
    \label{tab:ModelsAndEncodersOverviewAcc}
    \setlength\tabcolsep{2pt}
\begin{tabular}{llccccc}
\textbf{Dataset}&\textbf{Encoder}&\textbf{Model}&\textbf{Layer Size}&\textbf{\#L}&\textbf{Acc}&\textbf{AUC} \\ \hline
Cora (Node)&Llama2 13B&GIN&256&1&0.795&--\\
PubMed (Node)& E5&GCN&64&1&0.794&--\\
WikiCS (Node)&Llama2 13B&GAT&64&1&0.815&--\\[0.75ex]
WN18RR (Link)&Llama2 13B&GCN&128&1&0.725&--\\
Cora (Link)&E5&GAT&128&1&--&0.989\\
PubMed (Link)&E5&GAT&128&1&--&0.982\\[0.75ex]
BBBP (Graph)&Llama2 7B&GCN&64&1&--&0.648\\
BACE (Graph)&ST&GCN&64&3&--&0.767\\
\end{tabular}
\end{table}

The best expert-encoder pairs are shown for each dataset and task in Table~\ref{tab:ModelsAndEncodersOverviewAcc}.
For node classification, Llama2 13B and E5 have the best results.
In reasoning, the optimal combination is Llama2 13B with GCN, while for link prediction, the best pairing is E5 with a GAT. 
The best performance for graph classification is achieved with a GCN in combination with the sentence encoder or Llama2 7B. 
The GCN shows strong performance in graph classification and reasoning tasks, while the GAT is effective for graph classification and link prediction.

\begin{figure*}
     \centering
 \begin{subfigure}[b]{0.33\textwidth}
       \centering
  \includegraphics[width=1\textwidth]{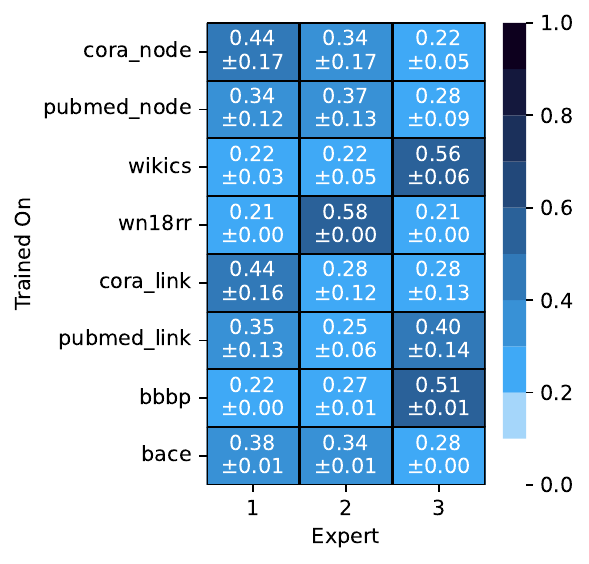}
  \caption{Cora (Node)}
    \label{fig:GateCoraNode}
   \end{subfigure} 
    \hfill
    \begin{subfigure}[b]{0.32\textwidth}
       \centering
  \includegraphics[width=1\textwidth]{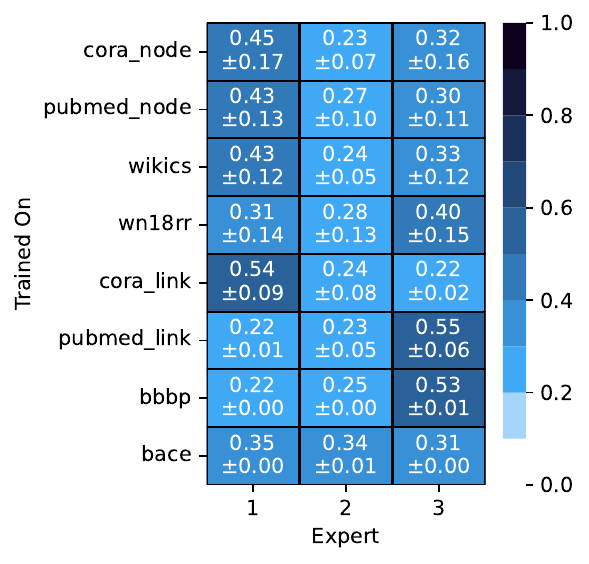}
  \caption{WN18RR}
    \label{fig:GateWN18RR}
   \end{subfigure} 
    \hfill
    \begin{subfigure}[b]{0.32\textwidth}
       \centering
  \includegraphics[width=1\textwidth]{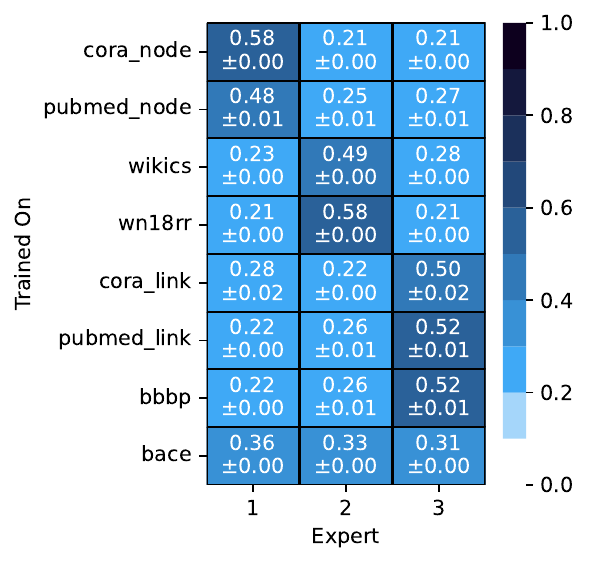}
  \caption{BACE}
    \label{fig:GateBACE}
   \end{subfigure} 
    \caption{The average scores for each expert on the test sets given the gating function of DA-MoE with $k=3$. The dataset for the test set is indicated in the sub-caption, and the gating function uses a multi-layer strategy.} 
        \label{fig:heatmapsGateDist}
\end{figure*}

\arxivonly{
\newcommand{\myscale}{0.22}
\begin{figure}
     \centering
 \begin{subfigure}[b]{\myscale\textwidth}
       \centering
  \includegraphics[width=1\textwidth]{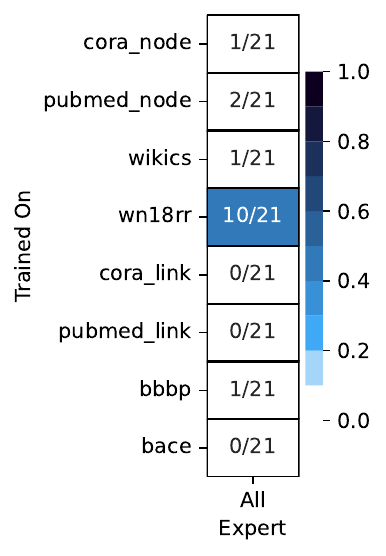}
  \caption{Cora (Node)}
    \label{fig:GateDisCoraNode}
   \end{subfigure} 
    \hfill
    \begin{subfigure}[b]{\myscale\textwidth}
       \centering
  \includegraphics[width=1\textwidth]{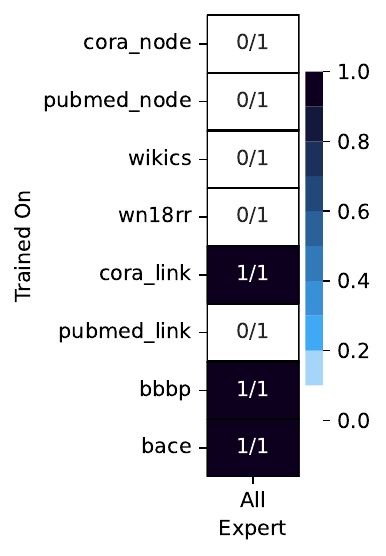}
  \caption{Cora (Link)}
    \label{fig:GateDisCoraLink}
   \end{subfigure} 

\begin{subfigure}[b]{\myscale\textwidth}
       \centering
  \includegraphics[width=1\textwidth]{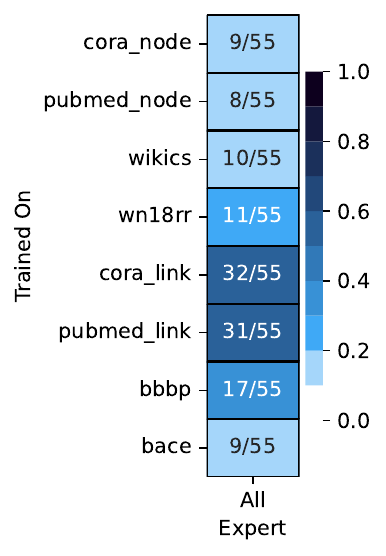}
  \caption{WN18RR}
    \label{fig:GateDisWN18RR}
   \end{subfigure} 
    \hfill
    \begin{subfigure}[b]{\myscale\textwidth}
       \centering
  \includegraphics[width=1\textwidth]{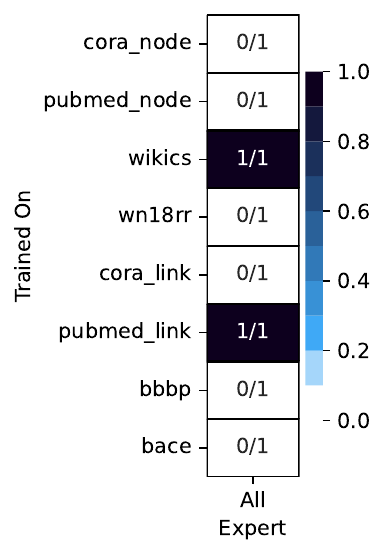}
  \caption{BACE}
    \label{fig:GateDisBACE}
   \end{subfigure} 
    \caption{The subfigures demonstrate the number of class pairs that exhibit a similar distribution of scores generated by the gating function. This gating function is part of the DA-MoE model with $k=3$ and uses a multi-layer strategy. The dataset for the test set is specified in the sub-caption.}
        \label{fig:fig4-heatmapsGate}
\end{figure}
}

The scores of the gating function of DA-MoE vary depending on the dataset and the strategy employed. Figure~\ref{fig:heatmapsGateDist} shows the average scores for each expert trained on the Cora (Node), WN18RR, and BACE datasets.
Although the average scores differ across training data and strategies, the three examples provide similar insights into their behavior.
Notably, the gating function trained on BACE demonstrates consistent behavior across all test sets, using each expert with a comparable frequency. Additionally, the gating function's performance varies depending on the specific test set employed.
Another observation is that the gating functions trained on Cora (Node) and Cora (Link) exhibit similar behaviors. 
Most classes show unique score distributions that differentiate them from others.
\arxivonly{The distribution of scores can be employed to distinguish between classes, as illustrated in Figure~\ref{fig:fig4-heatmapsGate}.} 

\section{Discussion}
\label{sec:discussion}

\arxivonly{\paragraph{Key Insights}}
In more than half of the cross-task or pre-training scenarios, Chimaera achieves performance close to that of models trained directly on the target dataset and task. 
Therefore, it shows strong generalization capabilities.
It also shows that Chimaera can use already trained GNNs and MoE (gating functions) for pre-training and cross-task scenarios.
For same-task and co-training, Chimaera works better with PRODIGY, outperforming linear GNNs as a classifier. 
However, when it comes to pre-training or cross-task scenarios, linear GNNs tend to be superior.
Overall, linear GNNs derive more benefits from the trained experts and gating mechanisms than PRODIGY. 
A possible explanation is that PRODIGY requires training for different tasks, whereas Chimaera effectively transfers knowledge across a single task and dataset. 
However, even with multi-task training, the Chimaera variant with PRODIGY is not better than using linear GNNs as a classifier in Chimaera.
Linear GNNs generally exhibit strong transfer capabilities across different tasks and datasets. 
Even when working with a limited number of examples, Chimaera enables the application of linear GNNs to various tasks while introducing new parameters that can be optimized. As a result, Chimaera enhances both the homogenization and generalizability of linear GNNs.
Additionally, our training method benefits linear GNNs.
The experiments demonstrate that training Chimaera as a whole reduces the transferability of linear GNNs. 
In other words, training Chimaera as a whole reduces the transferability of the experts and the gating functions.    
Thus, our training procedure, which consists of step-by-step training and freezing components, benefits Chimaera.

A Chimaera combining the experts with one of our strategies usually outperforms a Chimaera that uses no gating function and only one expert.
Moreover, combining different experts enhances performance across all tasks, except for graph-level tasks.
The score of a gating function depends on the training set, test set, and the strategy employed.  Additionally, we demonstrate that the gating functions exhibit varying behavior across most classes.
Using a larger number of experts tends to yield better results than using fewer. 
This is demonstrated by GMoE and DA-MoE, as $k=3$ performs better than $k=2$ for our MoE and, most of the time, also for our GFMs. 
Overall, the multi-layer strategy is the most effective for the MoE, as it typically leads to performance improvements. 
We recommend DA-MoE, as it uses the graph for gating and can be further tailored to specific needs.
The performance of experts with encoders using larger embeddings improves the results for datasets with a larger number of classes in a task.
This trend is consistent across all tasks we examined.  
An explanation is that larger embeddings make it easier to distinguish more classes.
Our findings suggest that the choice of encoder plays a crucial role in a GNN's performance. The size of the embeddings and the specific encoder selected strongly influence a GNN's effectiveness. 
Thus, selecting the right encoder is an important optimization consideration when applying GNNs to graphs.

\arxivonly{
\paragraph{Limitations}
\label{sec:limitations}
We only train our models for one task type.
This creates an issue with PRODIGY when it is not trained for a link-level task. In this case, the linear projection for the edge embeddings is not trained, and PRODIGY performs worse on link-level tasks.
However, to address this, we conduct experiments by training PRODIGY for multiple tasks.
Despite this effort, linear GNNs still outperform PRODIGY when dealing with unknown datasets.
Another limitation is that we do not train the experts, gating functions, and GFMs simultaneously without any pre-training. 
However, the training with linear GNN and unfrozen parameters does not improve performance; in fact, it appears to decrease it.
Additionally, testing all hyperparameter combinations of Chimaera significantly increases the time required for hyperparameter optimization when using grid search.
We use a maximum of three experts. However, even with this limit, we demonstrate that having more than one expert is advantageous. We leave the exploration of the optimal number of experts open for future work.
}

\section{Conclusion\arxivonly{ and Future Work}}
\label{sec:conclusion}
Chimaera is the first model to integrate gating functions and GFMs using multiple LLM embeddings.  
The combination of GNNs through gating functions enhances performance.
Additionally, using a gating function combined with MoE achieves better results for Chimaera than using the best expert.
Overall, Chimaera demonstrates strong generalization capabilities.
It enables linear GNNs to be applied to all classification tasks.
The application of linear GNNs facilitates transfer capabilities across different tasks and datasets, resulting in only a minimal decrease in performance compared to models specifically trained for those tasks.

\arxivonly{
Future research may explore additional strategies for combining experts and optimizing the gating functions (MoE). This may also include using a broader range of encoders, including those not based on LLMs, such as Node2Vec. Moreover, exploring different training procedures, including pre-training methods for GNNs and transductive GNNs, may lead to improved transfer performance. 

\section{Ethical Considerations}
Our paper presents foundational research and is not linked to specific applications.  While our work has broad implications, making GNN and similar models more applicable to broader training and testing settings, we believe that no specific societal consequences require immediate emphasis in this context.
}

\textbf{Acknowledgments}:
The authors acknowledge support by the state of Baden-Württemberg through bwHPC.
This work was co-funded by the Deutsche Forschungsgemeinschaft (DFG) as part of the
CodeInspector Project - 504226141.

\bibliographystyle{ieeetr}
\bibliography{references-short}

\arxivonly{
    \include{appendix}
}

\end{document}

%% file: appendix.tex
\clearpage

\appendix

\section{Comparison with GFT}

To compare Chimaera with other GFMs, we use the setup by~GFT~\cite{GFT}, which uses cross-domain and cross-task graph datasets.
It includes citation networks such as Cora, PubMed, arXiv, and the web link network WikiCS for node classification. 
Additionally, two Knowledge Graphs, WN18RR and FB15K237, are used for reasoning tasks.
For graph-level tasks, the molecule network HIV is used. 
The characteristics of the datasets and the specific training parameters are detailed in Table~\ref{tab:ComparisonsHyperparameters}.
Node embeddings are generated using Sentence-BERT~\cite{LLM:Sentence}, following the approach outlined by~GFT.  
All experiments are conducted ten times with different random seeds. For the HIV dataset, we evaluate using test AUC, while for the other datasets, we focus on accuracy.

For the comparison, we train a one-, two- and three-layer GCNs, all with a hidden size of $768$. Then, we combine these GCNs with a DA-MoE using a one-layer GCN with hidden size $768$ as a gating function. The gating function is also trained in the same manner as the GCNs.
During training, we use a weight decay of $0.001$, a drop rate of $0.2$, and the parameters shown in Table~\ref{tab:ComparisonsHyperparameters}. 
These parameters are based on the default settings used by GFT but have been slightly adjusted to accelerate the training process. For optimization, we use the AdamW optimizer~\cite{AdamW}.

\begin{figure}
     \centering

  \includegraphics[width=0.4\textwidth]{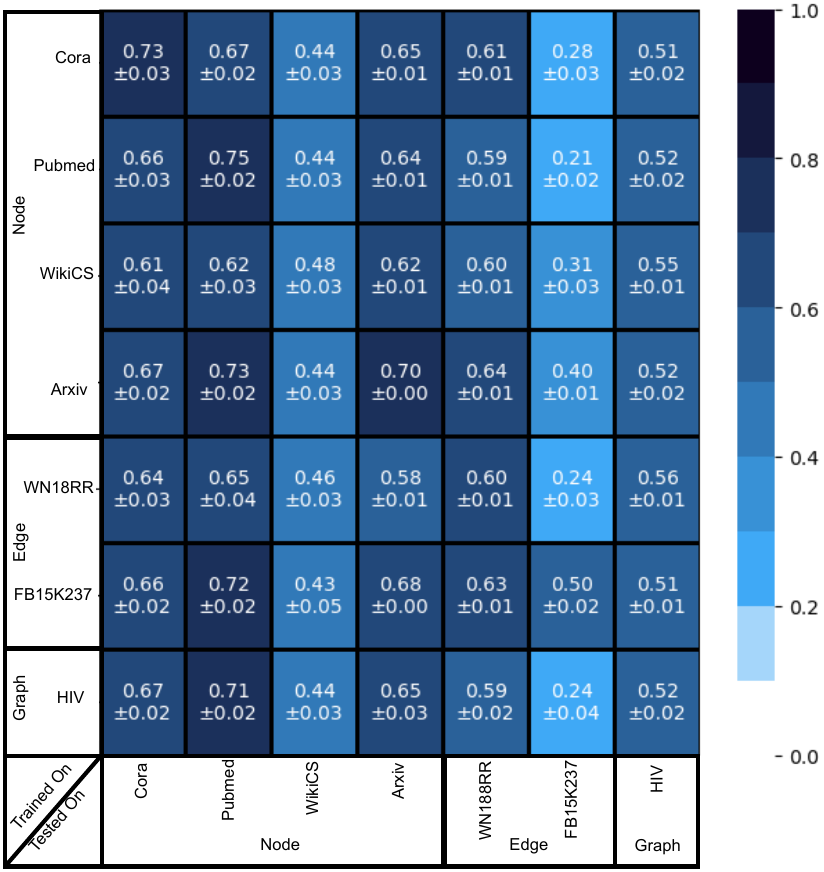}
    \caption{Chimaera's performance on the datasets of the comparison.}
        \label{fig:heatmapsChimeaeraLinearALl}
\end{figure}

For the classification tasks, we employ a linear GNN with different numbers of shots as the classifier for Chimaera. The experiment is conducted exclusively on Chimaera, while results for the other models are referenced from~GFT~\cite{GFT}.
In the first part,  we evaluate all trained Chimaera instances using the complete set of training examples for the linear GNNs.  
Figure~\ref{fig:heatmapsChimeaeraLinearALl} illustrates the performance of Chimaera across all datasets. It demonstrates that Chimaera effectively transfers knowledge between most datasets. 
Table~\ref{tab:NewCoTrainigSameTasks}  compares the performance of Chimaera and its components against other GFMs in the same-task and co-training setting.
Overall, Chimaera and its components demonstrate competitive performance compared to the other GFMs, showing promise for further development.
The performance decrease when using an untrained linear GNN instead of a trained classifier is typically less than five percentage points.

In the second part of our comparison, we evaluate Chimaera under various few-shot settings across different datasets.
These settings differ in the number of shots and the number of classes.
The tested classes are randomly chosen.
The results for the arXiv dataset are presented in Tables~\ref{tab:ArxivShotsPart1} and~\ref{tab:ArxivShotsPart2}. In the 3-way classification, Chimaera trained on FB15K237 outperforms all other baselines and Chimaeras. Overall, the Chimaera model trained on FB15K237 demonstrates strong performance on the arXiv dataset, surpassing models such as Prodigy and OFA. 
The results for the Cora dataset, shown in Table~\ref{tab:CoraShots}, indicate that the Chimaera trained on FB15K237 is the best-performing variant, again surpassing OFA. 
However, it is outperformed by GFT, which was adapted using additional fine-tuning instances. For the FB15K237 dataset, Table~\ref{tab:FB15K237Shots} shows that the Chimaera trained on arXiv achieves reasonable performance across all settings, even though it sometimes performs worse than the baselines. 
Regarding the WN18RR knowledge graph, the results can be found in Table \ref{tab:WN18RRShots}. Here, the Chimaera model trained on FB15K237 either matches or exceeds the performance of OFA. 
In the context of graph classification, the Chimaera demonstrates similar results for the HIV dataset, regardless of whether it was trained on this specific dataset, as indicated in Table~\ref{tab:HIVShots}.

In general, Chimaera achieves results that are similar to or better than those of OFA. Additionally, it can outperform GFT on arXiv when working with a smaller number of classes.
Our results show that Chimaera, which was trained on FB15K237, performs effectively on other datasets, demonstrating its ability to transfer knowledge between tasks and domains.  
Additionally, it can outperform models trained on multiple datasets.
Overall, Chimaera performs well on datasets with a small number of classes.
Furthermore, the high standard deviation in our results suggests that providing high-quality examples contributes positively to its performance. 
As a result, future improvements to Chimaera may also focus on selecting good examples and eliminating poor ones for the adaptation of the linear GNN.

\begin{table*}
    \centering
        \caption{The dataset characteristics (upper block) and training parameters (lower block) used for the comparison. ${}^*$\,Number of vertices and edges for HIV is the mean per graph.}\label{tab:ComparisonsHyperparameters}
    \setlength\tabcolsep{2pt}
    \begin{tabular}{l|cccc|cc|c}
& \multicolumn{4}{c|}{\textbf{Node Classification}}&\multicolumn{2}{c|}{\textbf{Reasoning}}&\textbf{Graph Classification} \\
  & Cora & PubMed & WikiCS & arXiv & WN18RR & FB15K237 &HIV\\\hline
 \textbf{Graphs }&1&1&1&1&1&1&41\,127\\
 \textbf{Vertices} &2\,708&19\,717&169\,343&11\,701&14\,541&40\,943&25.5${}^*$\\
  \textbf{Edges}&10\,556&44\,338&1\,166\,243&216\,123&310\,116&93\,003&27.5${}^*$\\
  \textbf{Classes}&7&3&40&10&237&11&2\\[2ex]
  \textbf{Epochs}&1\,000&1\,000&1\,000&2\,000&3\,000&3\,000&100\\
  \textbf{Batch size}&$-$&$-$&$-$&$-$&1\,024&1\,024&1\,024\\
  \textbf{Early stop}&200&200&200&500&200&200&20\\
  \textbf{Learning rate}&$0.005$&$0.005$&$0.0001$&$0.005$&$0.005$&$0.0001$&$0.005$
 \end{tabular}

\end{table*}

\begin{table*}
    \centering
        \caption{Model performance in pre-training and fine-tuning setting with SD}\label{tab:NewCoTrainigSameTasks}
    \setlength\tabcolsep{2pt}
\begin{tabular}{l|cccc|cc|c}
\multicolumn{1}{c|}{\textbf{Method}}& \multicolumn{4}{c|}{\textbf{Node Classification}}&\multicolumn{2}{c|}{\textbf{Reasoning}}&\textbf{Graph Classification} \\
& Cora & PubMed & WikiCS & arXiv & WN18RR & FB15K237 &HIV\\\hline
DGI~\cite{GNN:DGI}&$.721 $\small$\pm .003$&$.731 $\small$\pm .006$&$.753 $\small$\pm .010$&$.692 $\small$\pm .002$&$.758 $\small$\pm .006$&$.813 $\small$\pm .002$&$.596 $\small$\pm .012$\\
BGRL~\cite{BGRL}&$.712 $\small$\pm .003$&$.753 $\small$\pm .013$&$.765 $\small$\pm .007$&$.712 $\small$\pm .002$&$.754 $\small$\pm .003$&$.807 $\small$\pm .003$&$.640 $\small$\pm .011$\\
GraphMAE~\cite{GNN:GraphMAE}&$.731 $\small$\pm .004$&$.743 $\small$\pm .003$&$.776 $\small$\pm .004$&$.709 $\small$\pm .003$&$.790 $\small$\pm .005$&$.853 $\small$\pm .002$&$.610 $\small$\pm .006$\\
GIANT~\cite{GIANT}&$.751 $\small$\pm .005$&$.723 $\small$\pm .005$&$.766 $\small$\pm .009$&$.701 $\small$\pm .003$&$.844 $\small$\pm .003$&$.875 $\small$\pm .005$&$.654 $\small$\pm .014$\\
GFT~\cite{GFT}&\textbf{.786 \small$\pm$ .012}&\textbf{.772 \small$\pm$ .020}&\textbf{.794 \small$\pm$ .004}&$.719 $\small$\pm .001$&\textbf{.919 \small$\pm$ .003}&\textbf{.897 \small$\pm$ .002}&.\textbf{727 \small$\pm$ .014}\\[1ex]
1-layer GCN~\cite{GNN:GCN} &$.716$\small$\pm .023$&$.741$\small$\pm .019$&$.762$\small$\pm .005$&$.712$\small$\pm .002$&$.687$\small$\pm .005$&$.525$\small$\pm .013$&$.483$\small$\pm.020 $\\
2-layer GCN~\cite{GNN:GCN} &$.750 $\small$\pm .025$&$.745 $\small$\pm .025$&$.765 $\small$\pm .009$&$.722 $\small$\pm .003$&$.687 $\small$\pm .005$&$.466 $\small$\pm .067$&$.485 $\small$\pm .029$\\
3-layer GCN~\cite{GNN:GCN} &$.750 $\small$\pm .026$&$.748 $\small$\pm .023$&$.770 $\small$\pm .010$&$.695 $\small$\pm .069$&$.669 $\small$\pm .019$&$.386 $\small$\pm .049$&$.514 $\small$\pm .036$\\
DA-MoE~\cite{GNN:DA-MoE} &$.778 $\small$\pm .015$&$.769 $\small$\pm .022$&$.788 $\small$\pm .006$&\textbf{.728 \small$\pm$ .002}&$.691 $\small$\pm .003$&$.504 $\small$\pm .023$&$.543 $\small$\pm .033$\\
Chimaera &$.709 $\small$\pm .043$&$.739 $\small$\pm .030$&$.468 $\small$\pm .041$&$ .710$\small$\pm .003$&$.649 $\small$\pm .015$&$.503 $\small$\pm .026$&$.524 $\small$\pm .023 $
 \end{tabular}

\end{table*}

\begin{table*}
    \centering
        \caption{Few-shot learning performance on arXiv (Part 1). ``\# trains'' indicates the number of fine-tuning instances per class for GFT. The best performance is bold, and the best Chimaera is underlined.}\label{tab:ArxivShotsPart1}
    \setlength\tabcolsep{2pt}
    \begin{tabular}{l|ccc|ccc}
  \multicolumn{1}{c|}{\textbf{Method}}& \multicolumn{3}{c|}{\textbf{5-way}}&\multicolumn{3}{c}{\textbf{3-way}} \\
  & 5-shot & 3-shot & 1-shot & 5-shot & 3-shot & 1-shot \\\hline
 GPN~\cite{GPN} &$ .505$\small$\pm .031$ & $.483 $\small$\pm .038$&$.386 $\small$\pm .016$ &$.623 $\small$\pm .049$ &$.585 $\small$\pm .030$ &$.485 $\small$\pm .056$ \\
 TENT~\cite{TENT} &$ .608 $\small$\pm .075$ & $.560 $\small$\pm .089$&$.456 $\small$\pm .107 $ &$.742 $\small$\pm .099$ &$.705 $\small$\pm .115$ &$.594 $\small$\pm .136$ \\
 GLITTER~\cite{GLITTER} &$.560 $\small$\pm .044$ & $.574 $\small$\pm .049$&$.471 $\small$\pm .027$ &$.621 $\small$\pm .109$ &$.609 $\small$\pm .121$ &$.592 $\small$\pm .055$ \\
 TLP-BGRL~\cite{TLP} &$.501 $\small$\pm .088$ & $.462 $\small$\pm .079$&$.358 $\small$\pm .086$ &$.629 $\small$\pm .117$ &$.584 $\small$\pm .113$ &$.463 $\small$\pm .108$ \\
 TLP-SURGL~\cite{TLP}&\textbf{.779 \small$\pm$ .065 }& \textbf{.742 \small$\pm$ .076}&\textbf{.618\small$\pm$ .101} &$.863 $\small$\pm .075$ &$.838 $\small$\pm .089$ &$.735 $\small$\pm .127$ \\
 PRODIGY~\cite{GNN:PRODIGY}&$.611 $\small$\pm .059$ & $.586 $\small$\pm .058$&$.482 $\small$\pm .062$ &$.736 $\small$\pm .069$ &$.714 $\small$\pm .073$ &$.616 $\small$\pm .085$ \\
 OFA~\cite{GNN:OFA}&$.599 $\small$\pm .013$ & $.587 $\small$\pm .064$&$.528 $\small$\pm .039$ &$.722 $\small$\pm .033$ &$.718 $\small$\pm .016$ &$.605 $\small$\pm .027$ \\
 GFT (\# train = 5)~\cite{GFT}&$.680 $\small$\pm .019$ & $.660 $\small$\pm .025$&$.582 $\small$\pm .042$ &$.786 $\small$\pm .040$ &$.740 $\small$\pm .032$ &$.662 $\small$\pm .041$ \\
 GFT (\# train = 10)~\cite{GFT}&$.724 $\small$\pm .036$ & $.717 $\small$\pm .029$&$.624 $\small$\pm .026$ &$.788 $\small$\pm .012$ &$.762 $\small$\pm .042$ &$.699 $\small$\pm .038$ \\
 GFT (\# train = 20)~\cite{GFT}&$.731 $\small$\pm .028$ & $.717 $\small$\pm .024$&$.642 $\small$\pm .021$ &$.807 $\small$\pm .023$ &$.793 $\small$\pm .023$ &$.729 $\small$\pm .034$ \\
 GFT (\# train = 30)~\cite{GFT}&$.747 $\small$\pm .030$ & $.733 $\small$\pm .037$&$.651 $\small$\pm .038$ &$.796 $\small$\pm .026$ &$.768 $\small$\pm .027$ &$.732 $\small$\pm .034 $ \\[1ex]
  Chimaera (Cora)&$.628$\small $\pm .159$&$.592$\small $\pm .156$&$.469$\small $\pm .197$&$.746$\small $\pm .181$&$.787$\small $\pm .118$&$.743$\small $\pm .118$\\
  Chimaera (PubMed)&$.663$\small $\pm .191$&$.548$\small $\pm .178$&$.297$\small $\pm .183$&$.741$\small $\pm .192$&$.676$\small $\pm .182$&$.565$\small $\pm .173$\\
  Chimaera (WikiCS)&$.641$\small $\pm .150$&$.556$\small $\pm .204$&$.413$\small $\pm .192$&$.742$\small $\pm .146$&$.705$\small $\pm .180$&$.583$\small $\pm .229$\\
  Chimaera (WN18RR)&$.562$\small $\pm .143$&$.478$\small $\pm .194$&$.267$\small $\pm .134$&$.703$\small $\pm .134$&$.585$\small $\pm .179$&$.517$\small $\pm .214$\\
  Chimaera (FB15K237)&\underline{.762\small $\pm$ .127}&\underline{\textbf{.742\small $\pm$ .175}}&\underline{.586\small $\pm$ .163}&\underline{\textbf{.871\small $\pm$ .108}}&\underline{\textbf{.875\small $\pm$ .095}}&\underline{\textbf{.803\small $\pm$ .147}}\\
  Chimaera (HIV)&$.632$\small $\pm .189$&$.589$\small $\pm .188$&$.332$\small $\pm .199$&$.685$\small $\pm .224$&$.739$\small $\pm .160$&$.432$\small $\pm .224$
 \end{tabular}

\end{table*}

\begin{table*}
    \centering
        \caption{Few-shot learning performance on arXiv (Part 2). ``\# trains'' indicates the number of fine-tuning instances per class for GFT. The best performance is bold, and the best Chimaera is underlined.}\label{tab:ArxivShotsPart2}
    \setlength\tabcolsep{2pt}
    \begin{tabular}{l|ccc|ccc|ccc}
  \multicolumn{1}{c|}{\textbf{Method}}& \multicolumn{3}{c|}{\textbf{40-way}}&\multicolumn{3}{c|}{\textbf{20-way}}&\multicolumn{3}{c}{\textbf{10-way}}  \\
  & 5-shot & 3-shot & 1-shot & 5-shot & 3-shot & 1-shot & 5-shot & 3-shot & 1-shot  \\\hline
Prodigy~\cite{GNN:PRODIGY}&$.255 $\small $\pm .001$&$.237 $\small $\pm .001$&$.214 $\small $\pm .002$&$.343 $\small $\pm .004$&$.313 $\small $\pm.007 $&$.292 $\small $\pm .010$&$.508 $\small $\pm .018$&$.474 $\small $\pm .030$&$.411 $\small $\pm .063$\\
OFA~\cite{GNN:OFA}&$.240 $\small $\pm .006$&$.221 $\small $\pm .009$&$.213 $\small $\pm .013$&$.363 $\small $\pm .005$&$.326 $\small $\pm .002$&$.294 $\small $\pm .012$&$.496 $\small $\pm .027$&$.481 $\small $\pm .037$&$.395 $\small $\pm .054$\\
GFT (\# train = 5)~\cite{GFT}&$.363 $\small $\pm .010$&$.344 $\small $\pm .010$&$.265 $\small $\pm .011$&$.458 $\small $\pm .011$&$ .426$\small $\pm .012$&$.350 $\small $\pm .010$&$.564 $\small $\pm .035$&$ .524$\small $\pm .011$&$ .444$\small $\pm .026$\\
GFT (\# train = 10)~\cite{GFT}&$.418 $\small $\pm .009$&$.391 $\small $\pm .019$&$.308 $\small $\pm .006$&$.497 $\small $\pm .019$&$.469 $\small $\pm .015$&$.410 $\small $\pm .018$&$.602 $\small $\pm .011$&$.576 $\small $\pm .016$&$ .481$\small $\pm .025$\\
GFT (\# train = 20)~\cite{GFT}&$.451 $\small $\pm .012$&$.439 $\small $\pm .014$&$.350 $\small $\pm .015$&$.533 $\small $\pm .014$&$.509 $\small $\pm .017$&$.430 $\small $\pm .018$&$.635 $\small $\pm .016$&$.614 $\small $\pm .030$&$.532 $\small $\pm .018$\\
GFT (\# train = 30)~\cite{GFT}&\textbf{.467\small $\pm$ .011}&\textbf{.446\small $\pm$ .010}&\textbf{.359 \small $\pm$ .012}&\textbf{.540 \small $\pm$.014}&\textbf{.519 \small $\pm$ .012}&\textbf{.438 \small $\pm$ .019}&\textbf{.644 \small $\pm$ .013}&\textbf{.628 \small $\pm$ .09}&\textbf{ .547\small $\pm$ .016}\\[1ex]
Chimaera (Cora)&$.275$\small $\pm .065$&$.297$\small $\pm .047$&$.147$\small $\pm .074$&$.389$\small $\pm .115$&$.371$\small $\pm .078$&$.158$\small $\pm .114$&$.524$\small $\pm .144$&$.468$\small $\pm .139$&$.212$\small $\pm .154$\\
Chimaera (PubMed)&$.283$\small $\pm .052$&$.280$\small $\pm .063$&$.137$\small $\pm .070$&$.447$\small $\pm .124$&$.392$\small $\pm .106$&$.163$\small $\pm .132$&$.543$\small $\pm .147$&$.446$\small $\pm .150$&$.257$\small $\pm .172$\\
Chimaera (WikiCS)&$.285$\small $\pm .055$&$.274$\small $\pm .060$&$.123$\small $\pm .041$&$.411$\small $\pm .122$&$.357$\small $\pm .077$&$.170$\small $\pm .089$&$.493$\small $\pm .136$&$.393$\small $\pm .124$&$.239$\small $\pm .101$\\
Chimaera (WN18RR)&$.243$\small $\pm .069$&$.245$\small $\pm .048$&$.096$\small $\pm .047$&$.341$\small $\pm .169$&$.307$\small $\pm .104$&$.089$\small $\pm .067$&$.453$\small $\pm .166$&$.394$\small $\pm .141$&$.168$\small $\pm .095$\\
Chimaera (FB15K237)&\underline{.288\small $\pm$ .062}&\underline{.308\small $\pm$ .085}&\underline{.195\small $\pm$ .058}&\underline{.463\small $\pm$ .113}&\underline{.442\small $\pm$ .112}&\underline{.254\small $\pm$ .131}&\underline{.601\small $\pm$ .158}&\underline{.582\small $\pm$ .159}&\underline{.413\small $\pm$ .167}\\
Chimaera (HIV)&$.271$\small $\pm .097$&$.284$\small $\pm .062$&$.148$\small $\pm .058$&$.421$\small $\pm .160$&$.389$\small $\pm .113$&$.184$\small $\pm .121$&$.529$\small $\pm .176$&$.496$\small $\pm .167$&$.264$\small $\pm .190$
\end{tabular}

\end{table*}

\begin{table*}
    \centering
        \caption{Few-shot learning performance on Cora. ``\# trains'' indicates the number of fine-tuning instances per class for GFT. The best performance is bold and the best Chimaera is underlined.}\label{tab:CoraShots}
    \setlength\tabcolsep{2pt}
    \begin{tabular}{l|ccc|ccc|ccc}
  \multicolumn{1}{c|}{\textbf{Method}}& \multicolumn{3}{c|}{\textbf{7-way}}&\multicolumn{3}{c|}{\textbf{5-way}}&\multicolumn{3}{c}{\textbf{2-way}}  \\
  & 5-shot & 3-shot & 1-shot & 5-shot & 3-shot & 1-shot & 5-shot & 3-shot & 1-shot  \\\hline
GPN~\cite{GPN}&$-$&$-$&$-$&$-$&$-$&$-$&$.638 $\small $\pm .029$&$-$&$.561 $\small $\pm .021$\\
TENT~\cite{TENT}&$-$&$-$&$-$&$-$&$-$&$-$&$.590 $\small $\pm .024$&$-$&$.543 $\small $\pm .021$\\
 TLP-BGRL~\cite{TLP}&$-$&$-$&$-$&$-$&$-$&$-$&$.813 $\small $\pm .019$&$-$&$.592 $\small $\pm .025$\\
 TLP-SURGL~\cite{TLP}&$-$&$-$&$-$&$-$&$-$&$-$&\textbf{.925 \small $\pm$ .010}&$-$&$.815 $\small $\pm .021$\\[1ex]  
OFA~\cite{GNN:OFA}&$.321 $\small $\pm .018$&$.360 $\small $\pm .021$&$.304 $\small $\pm .024$&$.423 $\small $\pm .024$&$.313 $\small $\pm .026$&$.237 $\small $\pm .017$&$.722 $\small $\pm .038$&$.622 $\small $\pm .012$&$.519 $\small $\pm .044$\\
GFT (\# train = 1)~\cite{GFT}&$.436 $\small $\pm .074$&$.433 $\small $\pm .081$&$.414 $\small $\pm .080$&$.523 $\small $\pm .066$&$.515 $\small $\pm .066$&$.498 $\small $\pm .068$&$.750 $\small $\pm .041$&$.763 $\small $\pm .036$&$.0729 $\small $\pm .046$\\
GFT (\# train = 2)~\cite{GFT}&$.565 $\small $\pm .035$&$.559 $\small $\pm .035$&$.536 $\small $\pm .045$&$.637 $\small $\pm .034$&$.624 $\small $\pm .044$&$.605 $\small $\pm .046$&$.823 $\small $\pm .040$&$.817 $\small $\pm .038$&$.780 $\small $\pm .063$\\
GFT (\# train = 5)~\cite{GFT}&$.674 $\small $\pm .043$&$.673 $\small $\pm .044$&$.661 $\small $\pm .044$&$.741 $\small $\pm .043$&$.744 $\small $\pm .045$&$.727 $\small $\pm .049$&$.870 $\small $\pm .034$&$.860 $\small $\pm .033$&$.860 $\small $\pm .034$\\
GFT (\# train = 10)~\cite{GFT}&\textbf{.740 \small $\pm$ .039}&\textbf{.743 \small $\pm$ .037}&\textbf{.726 \small $\pm$ .038}&\textbf{.785 \small $\pm$ .030}&\textbf{.789 \small $\pm$ .026}&\textbf{.769 \small $\pm$ .022}&$.879 $\small $\pm .029$&\textbf{.885 \small $\pm$ .024}&\textbf{.884 \small $\pm$ .029}\\[1ex]
Chimaera (PubMed)&$.531$\small $\pm .044$&$.446$\small $\pm .043$&$.290$\small $\pm .048$&$.568$\small $\pm .090$&$.504$\small $\pm .068$&$.351$\small $\pm .065$&$.707$\small $\pm .083$&$.661$\small $\pm .102$&$.559$\small $\pm .103$\\
Chimaera (WikiCS)&$.437$\small $\pm .051$&$.335$\small $\pm .062$&$.270$\small $\pm .058$&$.478$\small $\pm .062$&$.402$\small $\pm .098$&$.335$\small $\pm .064$&$.712$\small $\pm .090$&$.669$\small $\pm .116$&$.657$\small $\pm .083$\\
Chimaera (arXiv)&$.580$\small $\pm .040$&$.540$\small $\pm .052$&$.422$\small $\pm .068$&$.633$\small $\pm .085$&$.612$\small $\pm .102$&$.480$\small $\pm .108$&$.831$\small $\pm .072$&$.793$\small $\pm .062$&$.691$\small $\pm .146$\\
Chimaera (WN18RR)&$.409$\small $\pm .045$&$.300$\small $\pm .062$&$.226$\small $\pm .039$&$.426$\small $\pm .052$&$.348$\small $\pm .101$&$.288$\small $\pm .028$&$.587$\small $\pm .105$&$.638$\small $\pm .078$&$.518$\small $\pm .116$\\
Chimaera (FB15K237)&\underline{.620\small $\pm$ .036}&\underline{.575\small $\pm$ .054}&\underline{.458\small $\pm$ .067}&\underline{.686\small $\pm$ .066}&\underline{.643\small $\pm$ .088}&\underline{.527\small $\pm$ .079}&\underline{.867\small $\pm$ .054}&\underline{.822\small $\pm$ .069}&\underline{.745\small $\pm$ .143}\\
Chimaera (HIV)&$.558$\small $\pm .056$&$.463$\small $\pm .084$&$.313$\small $\pm .070$&$.595$\small $\pm .097$&$.506$\small $\pm .133$&$.358$\small $\pm .112$&$.770$\small $\pm .095$&$.695$\small $\pm .115$&$.596$\small $\pm .151$
\end{tabular}

\end{table*}

\begin{table*}
    \centering
        \caption{Few-shot learning performance on FB15K237. ``\# trains'' indicates the number of fine-tuning instances per class for GFT. The best performance is bold and the best Chimaera is underlined.}\label{tab:FB15K237Shots}
    \setlength\tabcolsep{2pt}
    \begin{tabular}{l|ccc|ccc|ccc} 
  \multicolumn{1}{c!}{\textbf{Method}}& \multicolumn{3}{c|}{\textbf{40-way}}&\multicolumn{3}{c|}{\textbf{10-way}}&\multicolumn{3}{c}{\textbf{5-way}}  \\
  & 5-shot & 3-shot & 1-shot & 5-shot & 3-shot & 1-shot & 5-shot & 3-shot & 1-shot  \\\hline
Prodigy~\cite{GNN:PRODIGY}&$.620 $\small $\pm .006$&$.596 $\small $\pm .002$&$.543 $\small $\pm .007$&$.843 $\small $\pm .078$&$.796 $\small $\pm .083$&$.661 $\small $\pm .099$&$.881 $\small $\pm .007$&$.880 $\small $\pm .005$&$.876 $\small $\pm .008$\\
OFA~\cite{GNN:OFA}&$.665 $\small $\pm .003$&$.658 $\small $\pm .005$&$.635 $\small $\pm .009$&$.836 $\small $\pm .062$&$.831 $\small $\pm .015$&$.835 $\small $\pm .041$&$.914 $\small $\pm .006$&$.911 $\small $\pm .007$&$.910 $\small $\pm .010$\\
GFT (\# train = 10)~\cite{GFT}&$ .611$\small $\pm .016$&$.615 $\small $\pm .013$&$.608 $\small $\pm .014$&$.788 $\small $\pm .018$&$.791 $\small $\pm .016$&$.792 $\small $\pm .018$&$.863 $\small $\pm .011$&$.860 $\small $\pm .018$&$.877 $\small $\pm .009$\\
GFT (\# train = 20)~\cite{GFT}&$.704 $\small $\pm .017$&$.706 $\small $\pm .021$&$.702 $\small $\pm .014$&$.854 $\small $\pm .021$&$.856 $\small $\pm .013$&$.859 $\small $\pm .015$&$.918 $\small $\pm .011$&$.918 $\small $\pm .062$&$.918 $\small $\pm .015$\\
GFT (\# train = 30)~\cite{GFT}&\textbf{.750 \small $\pm$ .010}&\textbf{.746 \small $\pm$ .007}&\textbf{.750 \small $\pm$ .009}&\textbf{.891\small $\pm$ .017}&\textbf{.885 \small $\pm$ .022}&\textbf{.881 \small $\pm$ .014}&\textbf{.919 \small $\pm$ .01}&\textbf{.923 \small $\pm$ .019}&\textbf{.924 \small $\pm$ .013}\\[1ex]
Chimaera (Cora)&$.542$\small $\pm .148$&$.485$\small $\pm .157$&$.425$\small $\pm .158$&$.668$\small $\pm .128$&$.625$\small $\pm .160$&$.577$\small $\pm .174$&$.778$\small $\pm .064$&$.754$\small $\pm .120$&$.685$\small $\pm .161$\\
Chimaera (PubMed)&$.533$\small $\pm .135$&$.497$\small $\pm .135$&$.431$\small $\pm .159$&$.694$\small $\pm .143$&$.668$\small $\pm .145$&$.552$\small $\pm .220$&\underline{.817\small $\pm$ .097}&$.747$\small $\pm .146$&$.676$\small $\pm .173$\\
Chimaera (WikiCS)&$.546$\small $\pm .121$&$.485$\small $\pm .121$&$.394$\small $\pm .148$&$.628$\small $\pm .151$&$.609$\small $\pm .150$&$.568$\small $\pm .190$&$.693$\small $\pm .193$&$.709$\small $\pm .168$&$.676$\small $\pm .169$\\
Chimaera (arXiv)&$.520$\small $\pm .111$&$.489$\small $\pm .102$&\underline{.442\small $\pm$ .131}&\underline{.710\small $\pm$ .137}&\underline{.707\small $\pm$ .155}&\underline{.638\small $\pm$ .186}&$.816$\small $\pm .112$&\underline{.801\small $\pm$ .118}&\underline{.756\small $\pm$ .152}\\
Chimaera (WN18RR)&\underline{.556\small $\pm$ .107}&\underline{.499\small $\pm$ .123}&$.415$\small $\pm .136$&$.624$\small $\pm .141$&$.597$\small $\pm .153$&$.488$\small $\pm .148$&$.648$\small $\pm .135$&$.676$\small $\pm .129$&$.602$\small $\pm .173$\\
Chimaera (HIV)&$.488$\small $\pm .155$&$.472$\small $\pm .121$&$.406$\small $\pm .122$&$.697$\small $\pm .132$&$.682$\small $\pm .147$&$.590$\small $\pm .190$&$.786$\small $\pm .075$&$.781$\small $\pm .113$&$.625$\small $\pm .148$
\end{tabular}
\end{table*}

\begin{table*}
    \centering
        \caption{Few-shot learning performance on WN18RR. ``\# trains'' indicates the number of fine-tuning instances per class for GFT. The best performance is bold and the best Chimaera is underlined.}\label{tab:WN18RRShots}
    \setlength\tabcolsep{2pt}
    \begin{tabular}{l|ccc|ccc|ccc}
  \multicolumn{1}{c|}{\textbf{\shortstack[l]{Method}}}& \multicolumn{3}{c|}{\textbf{10-way}}&\multicolumn{3}{c|}{\textbf{5-way}}&\multicolumn{3}{c}{\textbf{3-way}}  \\
  & 5-shot & 3-shot & 1-shot & 5-shot & 3-shot & 1-shot & 5-shot & 3-shot & 1-shot  \\\hline
OFA~\cite{GNN:OFA}&$.326 $\small $\pm .016$&$.306 $\small $\pm .010$&$.258 $\small $\pm .011$&$.483 $\small $\pm .032$&$.450 $\small $\pm .024$&$.344 $\small $\pm .015$&$.607 $\small $\pm .038$&$.613 $\small $\pm .026$&$.518 $\small $\pm .027$\\
GFT (\# train = 1)~\cite{GFT}&$.355 $\small $\pm .046$&$.355 $\small $\pm .050$&$.353 $\small $\pm .042$&$.488 $\small $\pm .036$&$.485 $\small $\pm .037$&$.481 $\small $\pm .044$&$.626 $\small $\pm .027$&$.607 $\small $\pm .039$&$.584 $\small $\pm .038$\\
GFT (\# train = 2)~\cite{GFT}&$.424 $\small $\pm .031$&$.425 $\small $\pm .029$&$.420 $\small $\pm .030$&$.559 $\small $\pm .026$&$.548 $\small $\pm .023$&$.544 $\small $\pm .022$&$ .663$\small $\pm .019$&$ .664$\small $\pm .017$&$.649 $\small $\pm .033$\\
GFT (\# train = 5)~\cite{GFT}&$.448$\small $\pm .029$&$.449 $\small $\pm .031$&$.448 $\small $\pm .035$&$.580 $\small $\pm .026$&$.577 $\small $\pm .022$&$.574 $\small $\pm .025$&$ .689$\small $\pm .022$&$.697 $\small $\pm .021$&$ .688$\small $\pm .010$\\
GFT (\# train = 10)~\cite{GFT}&\textbf{.512\small $\pm$ .046}&\textbf{.513 \small $\pm$ .048}&\textbf{.509 \small $\pm$ .042}&\textbf{.627 \small $\pm$ .049}&\textbf{.630 \small $\pm$ .059}&\textbf{.636 \small $\pm$ .063}&$.722$\small $\pm .037$&\textbf{.726 \small $\pm$ .049}&\textbf{.726 \small $\pm$ .048}\\[1ex]
Chimaera (Cora)&$.296$\small $\pm .048$&$.259$\small $\pm .026$&$.190$\small $\pm .050$&$.455$\small $\pm .117$&$.424$\small $\pm .107$&$.366$\small $\pm .116$&$.599$\small $\pm .155$&$.625$\small $\pm .126$&$.479$\small $\pm .172$\\
Chimaera (PubMed)&$.299$\small $\pm .046$&$.274$\small $\pm .062$&$.208$\small $\pm .046$&$.478$\small $\pm .131$&$.386$\small $\pm .066$&$.293$\small $\pm .101$&$.614$\small $\pm .151$&$.550$\small $\pm .120$&$.404$\small $\pm .124$\\
Chimaera (WikiCS)&$.290$\small $\pm .045$&$.249$\small $\pm .030$&$.186$\small $\pm .046$&$.442$\small $\pm .133$&$.362$\small $\pm .101$&$.295$\small $\pm .069$&$.573$\small $\pm .125$&$.514$\small $\pm .117$&$.458$\small $\pm .151$\\
Chimaera (arXiv)&$.307$\small $\pm .038$&$.313$\small $\pm .044$&$.243$\small $\pm .078$&$.518$\small $\pm .144$&$.514$\small $\pm .146$&$.388$\small $\pm .127$&$.711$\small $\pm .154$&$.690$\small $\pm .158$&$.531$\small $\pm .197$\\
Chimaera (FB15K237)&\underline{.312\small $\pm$ .036}&\underline{.317\small $\pm$ .038}&\underline{.256\small $\pm$ .064}&\underline{.533\small $\pm$ .147}&\underline{.530\small $\pm$ .146}&\underline{.429\small $\pm$ .132}&\underline{\textbf{.729\small $\pm$ .161}}&\underline{.720\small $\pm$ .151}&\underline{.596\small $\pm$ .189}\\
Chimaera (HIV)&$.308$\small $\pm .048$&$.296$\small $\pm .070$&$.210$\small $\pm .097$&$.492$\small $\pm .140$&$.440$\small $\pm .128$&$.337$\small $\pm .175$&$.644$\small $\pm .172$&$.624$\small $\pm .151$&$.563$\small $\pm .165$
\end{tabular}
\end{table*}

\begin{table*}
    \centering
        \caption{Few-shot learning performance on HIV. ``\# trains'' indicates the number of fine-tuning instances per class for GFT. The best performance is bold and the best Chimaera is underlined.}\label{tab:HIVShots}
    \setlength\tabcolsep{2pt}
    \begin{tabular}{l|cccc} 
  \multicolumn{1}{c|}{\textbf{Method}}& \multicolumn{4}{c|}{\textbf{2-Way}}  \\
  &10-shot & 5-shot & 3-shot & 1-shot  \\\hline
OFA~\cite{GNN:OFA}&$.544 $\small $\pm .049$&$.576 $\small $\pm .037$&\textbf{.593 \small $\pm$ .030}&$.572 $\small $\pm .018$\\
GFT (\# train = 10)~\cite{GFT}&$.532 $\small $\pm .118$&$.542 $\small $\pm .106$&$.576 $\small $\pm .105$&$.583 $\small $\pm .091$\\
GFT (\# train = 20)~\cite{GFT}&\textbf{.587 \small $\pm$ .075}&\textbf{.588 \small $\pm$ .069}&$.584 $\small $\pm .073$&\textbf{.599 \small $\pm$ .071}\\
GFT (\# train = 30)~\cite{GFT}&$.581 $\small $\pm .053$&$.586 $\small $\pm .051$&$.583 $\small $\pm .054$&$.591 $\small $\pm .052$\\[1ex]
Chimaera (Cora)&$.506$\small $\pm .024$&$.514$\small $\pm .027$&$.512$\small $\pm .024$&$.512$\small $\pm .023$\\
Chimaera (PubMed)&\underline{.525\small $\pm$ .037}&$.522$\small $\pm .037$&$.523$\small $\pm .036$&$.524$\small $\pm .037$\\
Chimaera (WikiCS)&$.500$\small $\pm .030$&$.495$\small $\pm .026$&$.494$\small $\pm .026$&$.500$\small $\pm .029$\\
Chimaera (arXiv)&$.524$\small $\pm .022$&\underline{.528\small $\pm$ .022}&\underline{.526\small $\pm$ .021}&\underline{.529\small $\pm$ .023}\\
Chimaera (WN18RR)&$.524$\small $\pm .031$&$.526$\small $\pm .032$&$.525$\small $\pm .031$&$.527$\small $\pm .033$\\
Chimaera (FB15K237)&$.518$\small $\pm .022$&$.522$\small $\pm .025$&$.518$\small $\pm .022$&$.522$\small $\pm .025$\end{tabular}
\end{table*}